\documentclass[letterpaper]{article} 
\usepackage{aaai2026}  
\usepackage{times}  
\usepackage{helvet}  
\usepackage{courier}  
\usepackage[hyphens]{url}  
\usepackage{graphicx} 
\usepackage{natbib}  
\usepackage{caption} 
\usepackage{algorithm}
\usepackage{algorithmic}
\usepackage{subfig}
\usepackage[colorlinks=true, linkcolor=blue, urlcolor=blue, citecolor=black]{hyperref}

\usepackage{newfloat}
\usepackage{listings}
\usepackage{authblk}
\DeclareCaptionStyle{ruled}{labelfont=normalfont,labelsep=colon,strut=off} 
\floatstyle{ruled}
\newfloat{listing}{tb}{lst}{}
\floatname{listing}{Listing}
\usepackage{amsmath}
\usepackage{amsfonts}
\usepackage{multirow}
\usepackage{comment}

\title{DiLO: Disentangled Latent Optimization for Learning Shape and Deformation in Grouped Deforming 3D Objects}
\author[]{Mostofa Rafid Uddin$^{1}$, Jana Armouti$^{1}$, Umong Sain$^{2}$, Md Asib Rahman$^{2}$, Xingjian Li$^{1}$, Min Xu$^{1,\dagger}$

\vspace{2em}

$^{1}$ Carnegie Mellon University, Pittsburgh, PA 15213, USA.

$^{2}$ Bangladesh University of Engineering and Technology, Dhaka 1000, Bangladesh

\vspace{2em}

$^{\dagger}$Corresponding Author: Min Xu (mxu1@cs.cmu.edu)}

\begin{document}

\maketitle

\begin{abstract}
In this work, we propose a disentangled latent optimization-based method for parameterizing grouped deforming 3D objects into shape and deformation factors in an unsupervised manner. Our approach involves the joint optimization of a generator network along with the shape and deformation factors, supported by specific regularization techniques. For efficient amortized inference of disentangled shape and deformation codes, we train two order-invariant PoinNet-based encoder networks in the second stage of our method. We demonstrate several significant downstream applications of our method, including unsupervised deformation transfer, deformation classification, and explainability analysis. Extensive experiments conducted on 3D human, animal, and facial expression datasets demonstrate that our simple approach is highly effective in these downstream tasks, comparable or superior to existing methods with much higher complexity.
\end{abstract}

\section{Introduction}

Parameterizing 3D objects with distinct generative factors, such as shape and deformation, has garnered considerable attention in computer graphics and vision research \cite{cosmo2020limp, chen2021intrinsic, huang2021arapreg, aumentado2019geometric}. In this context, shape typically refers to the intrinsic properties of 3D objects, such as height, body structure, and surface geometry, while deformation pertains to extrinsic properties, including pose, motion, twisting, and morphing. By disentangling these generative factors and parameterizing 3D objects accordingly, it is possible to achieve efficient 3D deformation transfer, shape manipulation, and generation\cite{zhou2020unsupervised, song2023unsupervised, sun2023mapconnet, cosmo2020limp, aumentado2019geometric}. This capability has practical applications in industries such as content creation, gaming, and AR/VR \cite{chen2021aniformer, chen2023smg}. 

For nearly a decade, parameterizing specific 3D objects, such as humans \cite{anguelov2005scape, loper2023smpl, pons2015dyna}, hands \cite{romero2022embodied}, and faces \cite{li2017learning, ploumpis2019combining}, into generative factors like shape and deformation has been achieved using hand-crafted features, such as, landmarks, skeletons, or manually estimated point-wise distances. However, obtaining such features requires significant manual effort and expert knowledge. Additionally, for many deforming 3D objects (e.g., organs, proteins), it is often impractical to define clear skeletons or landmark features. Addressing these limitations, recent advancements have led to the development of high-fidelity deep representation learning models \cite{chen2021intrinsic, zhou2020unsupervised, cosmo2020limp, aumentado2019geometric} that can parameterize 3D objects into distinct generative factors- shape and deformation codes in an unsupervised manner, eliminating the need for manually provided features. 
\footnote{often referred to as pose, we use the term deformation since our method can disentangle both pose and non-pose deformations}
\begin{figure*}[!ht]
    \centering
    \includegraphics[width=0.9\linewidth, height=0.35\linewidth]{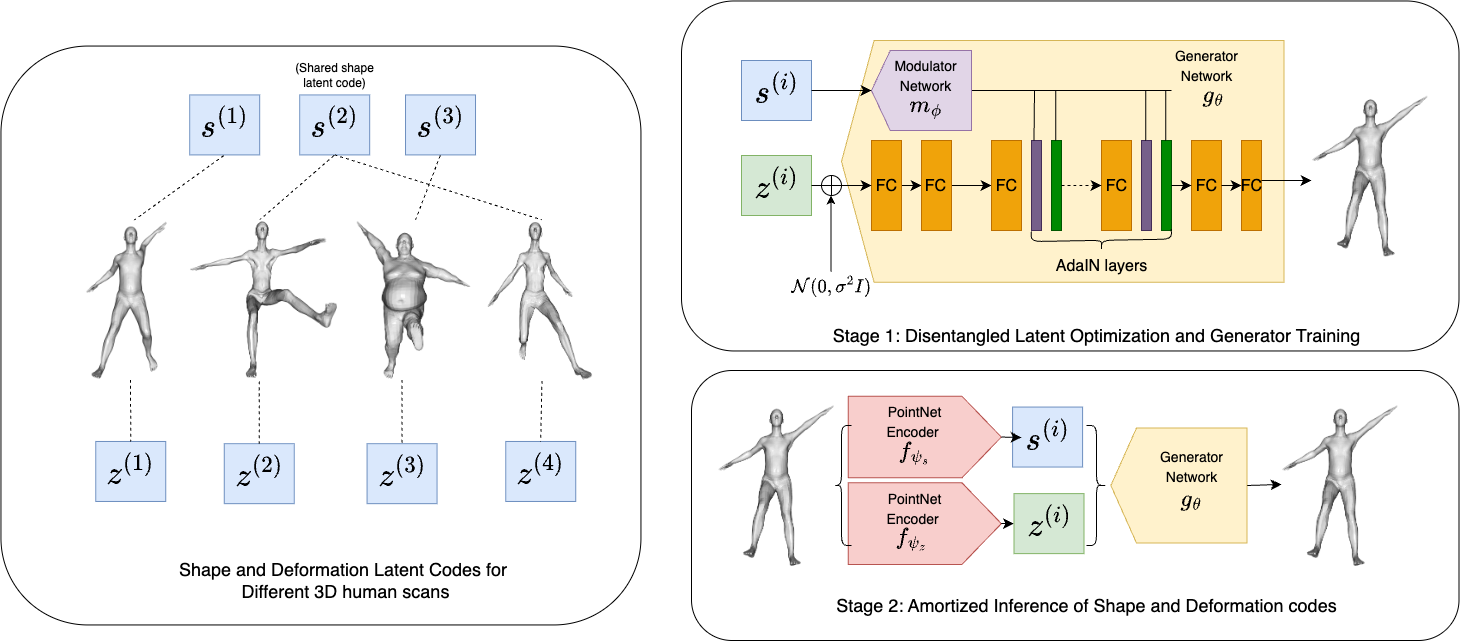}
    \caption{\textbf{An overview of our proposed unsupervised shape deformation disentanglement method.} On the left, we show the conceptualization of shape codes $s^{(i)}$ and deformation codes $z^{(i)}$. On the right, we demonstrate the two learning stages of our method. In stage 1, we optimize $s^{(i)}$ and $z^{(i)}$ together with a generator network. In stage 2, we infer the optimized codes $s^{(i)}$ and $z^{(i)}$ from the input 3D object using two PointNet \cite{qi2017pointnet} encoders.}
    \label{fig:method}
\end{figure*}

These deep learning-based methods are generally trained on grouped 3D object collections, where multiple deforming 3D objects are grouped based on their shape, deformation, or other characteristics. The methods leverage group information of their shapes and utilize assumptions specific to deformation to train their models. For example, \cite{cosmo2020limp} enforces that 3D objects with two different deformations of the same shape preserve the geodesic distance between their vertices. \cite{zhou2020unsupervised} model deformation in the 3D objects as As-Rigid-As-Possible (ARAP) deformation and enforces it during model training. Despite some success, these methods still leave much room for improvement. Moreover, implementing the deformation-specific constraints during training creates much computational overhead and is thus resource-intensive.

In this work, we approach the problem from a different perspective. We leverage the grouping information of the shapes of the 3D objects to learn two generative factors, one responsible for the commonalities within each group, in other words, shape, and the other accountable for intra-group instance-wise variation, which, in our datasets, is deformation. To this end, we developed a novel method, called \textbf{disentangled latent optimization (DiLO)}. DiLO is a two-stage framework. In the first stage,  we perform latent optimization of the generative factors using an autodecoder network \cite{huang2021arapreg}. Instead of optimizing individual latent factors for both shape and deformation for each 3D object, we perform \textbf{shared optimization for the shape factors} since they are responsible for commonalities within the group. In this way, all objects belonging to a shape group have the same shape factor, whereas they have different deformation factors. This design ensures disentanglement between the shape and deformation codes. For training the generator of the autodecoder network, we incorporate adaptive instance normalization (AdaIn) layers commonly used for image style transfer \cite{gabbay2021scaling, gabbaydemystifying}. Whereas the deformation code is directly passed as input to the generator, the shape code is used to predict the parameters of the AdaIn layers of the generator. In the second stage, we train two permutation-invariant PointNet encoder \cite{qi2017pointnet} networks that can infer the optimized latent codes from any given 3D object, enabling fast amortized inference. Our framework is lightweight, easy to train, and does not suffer from training instability, unlike many methods that rely on adversarial training.

We conducted extensive experiments to evaluate our method and baseline approaches across multiple 3D datasets, specifically 3D human models (SMPL \cite{loper2023smpl}), 3D animal models (SMAL \cite{zuffi20173d}), and 3D face models (COMA \cite{ranjan2018generating}). The evaluations were performed on three primary tasks: 1) unsupervised 3D deformation transfer, 2) deformation classification from latent codes, and 3) explainability analysis. Both qualitative and quantitative results consistently demonstrated the efficacy and superiority of our method compared to existing baselines. Additionally, we conducted extensive ablation experiments that demonstrate the effectiveness of individual components in our method. We summarize our major contributions as follows:
\begin{itemize}
    \item We introduce \textbf{disentangled} latent optimization to learn shape and deformation for grouped deforming 3D shapes in an unsupervised manner. 
    \item We demonstrate multiple downstream applications of our method, including unsupervised 3D deformation transfer, deformation classification, and explainability analyses using benchmark 3D shape datasets.
    \item Our method is computationally efficient, practically explainable, and also offers strong performance compared to the complex baseline methods in downstream tasks. 
\end{itemize}


\section{Related Works}
\label{sec:related_works}
\textbf{Unsupervised shape deformation disentanglement: }Parameterizing 3D shapes into shapes and deformation-specific components has been studied in computer graphics and vision for quite some time. Earlier methods used variants of principal component analysis to disentangle the shape and deformation-specific factors \cite{rustamov2013map, corman2017functional, gao2017data}. Such disentanglement works when shape and deformation are linearly separable but not when non-linearity is present. Tan et al. \cite{tan2018variational} applied variational autoencoders (VAE) on 3D data and captured factors that are not linearly separable across different dimensions of the VAE latent code. Nevertheless, there was still no explicit disentanglement of shape and deformation-specific factors. GD-VAE \cite{aumentado2019geometric} was among the first methods to perform explicit disentanglement of shape and deformation factors into two different latent codes. However, they did not exploit the shape group information of the datasets. Subsequently, several methods \cite{chen2021intrinsic, zhou2020unsupervised, cosmo2020limp} based on VAEs or GANs were developed to disentangle shape and deformation by leveraging shape group information. These approaches consistently outperformed GD-VAE, highlighting the effectiveness of using group information. DiLO also utilizes shape group information for disentanglement; however, unlike prior methods, it avoids computationally expensive operations such as geodesic distance calculations or ARAP deformation, achieving efficient disentanglement without sacrificing performance.

\textbf{Unsupervised 3D deformation transfer: }3D deformation transfer is a major downstream application of our unsupervised 3D shape-deformation disentanglement method. The deformation transfer task aims to transfer the deformation of one 3D object into another while keeping the same shape or identity. Deformation transfer methods \cite{song2023unsupervised, song20213d} directly infer the deformation transferred object. Most existing deformation transfer methods \cite{song20213d, wang2020neural, sumner2004deformation} are supervised, using the ``group truth" transferred mesh as the target. Very recently, a few unsupervised 3D deformation transfer methods \cite{song2023unsupervised, sun2023mapconnet} have been developed. X-DualNet \cite{song2023unsupervised} uses dual reconstruction and consistency losses similar to \cite{zhou2020unsupervised} to perform unsupervised deformation transfer. MAPConNet \cite{sun2023mapconnet}, uses mesh-level and point-level contrastive learning. Our shape-deformation disentanglement method can also be used for the unsupervised deformation transfer through latent manipulation and 3D generation (details on Section \ref{sec: pose_transfer}). However, unlike these methods, our method is generative and capable of various 3D shape analysis tasks through latent space manipulation and 3D generation.

Further discussions on several other related works can be found in the supplementary document.

\section{Method}
\label{sec:method}
Given a set of $N$ deforming 3D objects ${\{x^{(i)}\}}_{i=1}^N$ and their shape group information, the goal of our method is to learn disentangled latent codes for shape and deformation-specific information. Simultaneously, our method aims to learn a generator that allows controllable generation of 3D objects. Considering a 3D point-cloud or mesh input space $X$, shape space $S$, and deformation space $Z$ disentangled from $S$, our method learns the spaces $S$ and $Z$ as well as a mapping $g_\theta: S\times Z \rightarrow X$. 

For any \( (i, j) \), with \( s^{(i)}, s^{(j)} \in S \) and \( z^{(i)}, z^{(j)} \in Z \), the outputs \( g_\theta(s^{(i)}, z^{(i)}) \) and \( g_\theta(s^{(j)}, z^{(j)}) \) should satisfy:
\[
\begin{cases}
s^{(i)} = s^{(j)}, \, z^{(i)} \neq z^{(j)} & \Leftrightarrow \text{same shape, distinct deform.}, \\
s^{(i)} \neq s^{(j)}, \, z^{(i)} = z^{(j)} & \Leftrightarrow \text{same deform., distinct shape}.
\end{cases}
\]

\subsection{Overview}
At a high level, our method is built on an auto-decoder \cite{huang2021arapreg} based architecture  (Figure \ref{fig:method}). For each point-cloud or mesh $x^{(i)}$ in the input space $X \subseteq \mathbb{R}^{N\times V \times 3}$, we learn a shape latent code $s^{(i)}$ in the shape space $S \subseteq \mathbb{R}^{N\times d_s}$ and deformation latent code $z^{(i)}$ in the deformation space $Z \subseteq \mathbb{R}^{N\times d_z}$, where $d_s$ and $d_z$ are dimensions of shape code and deformation code respectively and $V$ is the number of points in each $x^{(i)}$. Simultaneously, we learn the mapping $g_\theta$ as a decoder or generator network. 

\subsection{Disentanglement of shape and deformation}

Merely optimizing two latent codes with the autodecoder network does not result in the disentanglement of the latent codes. Specialized techniques are needed to ensure the shape code represents only shape information, and the deformation code represents only deformation information. As mentioned below, this is achieved by processing the shape and deformation codes differently. 

\subsubsection{Shape Code Optimization}
If two 3D objects $x^{(i)}$ and $x^{(j)}$ have the same shape, their shape codes $s^{(i)}$ and $s^{(j)}$ should also be the same. This constraint is enforced through the shape group information. If two 3D objects belong to the same shape group, they are assigned the same shape identity label (Figure \ref{fig:method}). While optimizing the shape latent code, we explicitly constrain the shape latent codes to be shared within the same shape group of 3D objects. Such explicit constraint makes it difficult to have any deformation information in the shape code. Additionally, optimizing the shape latent codes for each shape group directly, rather than inferring the shape code for every 3D object and then averaging the codes for each group to create a template, allows us to simply use random sampling for creating mini-batches during training.

\subsubsection{Deformation Code Optimization}
For each 3D object $x^{(i)} \in X$ in the dataset, we optimize a deformation code $z^{(i)} \in Z$. Since information represented by the deformation code should be minimal and not exhibit shape-specific attributes, the code needs to be regularized. To this end, we use the following two ways to regularize the deformation codes.

First, we optimize the deformation latent codes with L2 regularization. Such regularization encourages the values of the deformation latent codes to be small and close to zero. Second, we add a Gaussian noise of zero mean and fixed variance to the deformation codes before passing them to the generator network $g_\theta$. This is unlike variational autoencoders (VAE), where the variances are learned. This mechanism ensures the variance does not decrease to a very small value for any particular component and thus prevents partial posterior collapse \cite{}. 

\subsubsection{Generator Network Optimization}
Our generator network $g_\theta$ consists of multiple fully connected linear layers with several adaptive instance normalization (AdaIN) layers. During training, we pass the deformation code with additive noise $z^{(i)} + \epsilon, \epsilon \sim \mathcal{N}(0, \sigma^{2} I)$ directly as input to the generator. However, for shape latent code $s^{(i)}$, we do not directly pass it to the generator $ g_\theta$. Instead, we use it to predict the parameter values of the adaptive instance normalization (AdaIN) layers in $g_\theta$. We achieve this by passing the shape code to a modulator network $m_\phi$ that predicts the parameter values for the AdaIn layers in $g_\theta$. The parameters of AdaIn layers are used to scale and shift the input features. 

Assume there are $J$ AdaIn layers, the parameters of the $j$-th AdaIn layer are $\gamma_j$ and $\beta_j$, and input to the $j$-th AdaIn layer is $z_j'^{(i)}$ where $z^{(i)}$ is used as input to generator $g_\theta$, then the output of the $j$-th AdaIn layer will be, 
\begin{align}
    o_{j} = \gamma_j \times \frac{z_j'^{(i)} - \mu (z^{(i)})}{\sigma (z_j'^{(i)})} + \beta_j, \texttt{    for } j = \{1, 2, \dots, J\}
\end{align}
\begin{align}
    \{\gamma_j, \beta_j\}_{j=1}^J = m_\phi(s^{(i)})
\end{align}
The output $o_{j}$ passed through a fully connected linear layer $l^j$ is used as the input to the $(j+1)$-th AdaIN layer.
\begin{align}
    z_{j+1}'^{(i)} = l^j (o_j), \texttt{    for } j = \{1, 2, \dots, J\}
\end{align}
The output of the $J$-th AdaIn layer $z_{J+1}'^{(i)}$ is passed through a sequence of linear layers and activation functions which finally generates the output point cloud $y^{(i)}$ of size $\mathbb{R}^{N \times 3}$. The similarity between the output point cloud $y^{(i)}$ and the point cloud $x^{(i)}$ is maximized using a reconstruction loss $L_{\text{recon}}$. 

The ouput of the final AdaIn layer is passed through a sequence of linear layers and activation functions which finally generates the output point cloud $y^{(i)}$ of size $\mathbb{R}^{N \times 3}$. The similarity between the output point cloud $y^{(i)}$ and the point cloud $x^{(i)}$ is maximized using a reconstruction loss $L_{\text{recon}}$. Overall, the loss function $L1$ of our method at this stage becomes:
\begin{align}
\label{eq: latent}
    L1 =& L_{\text{recon}}(y^{(i)}, x^{(i)}) + \lambda \| z^{(i)} \|_2^2\\
      =& L_{\text{recon}} (g_\theta(z^{(i)}+\epsilon, s^{(i)}), x^{(i)}) + \lambda \| z^{(i)} \|_2^2, \epsilon \sim \mathcal{N}(0, \sigma^{2} I)
\end{align}
We optimize $\{z^{(i)}, s^{(i)}\}_{i=1}^N$ along with the parameters $\theta$ in $g_\theta$ and $\phi$ in $m_\phi$, all using the loss function $L1$ and gradient descent. 

\subsection{Inference of Shape and Deformation Codes}
In the above steps, we optimize the shape and deformation codes for each 3D object in the input dataset. However, we should be able to infer the shape and deformation codes for any point cloud or mesh not present in the input dataset. To this end, in the second stage, we learn two inverse mappings $f_{\psi_s}: X \rightarrow S$ and $f_{\psi_z}: X \rightarrow Z$ with two encoder networks. Given a point-cloud or mesh as input $x^{(i)}$, the encoder $f_{\psi_s}$ outputs the corresponding shape code $s^{(i)}$ and the encoder $f_{\psi_z}$ outputs the corresponding deformation code $z^{(i)}$ that has been learned in the first stage. We ensure this with a distance loss $L_\text{dis}$ between the encoder predictions and the latent codes learned in the first stage. We also use a reconstruction loss to ensure that the encoder-predicted latent codes can be used to reconstruct $x^{(i)}$.
Overall, the loss function $L2$ in the second stage of our method becomes:
\begin{align}
\label{eq: amortized}
    L2 =& L_{\text{recon}}(g_\theta(f_{\psi_z}(x^{(i)}), f_{\psi_s}(x^{(i)})), x^{(i)}) \\
    &+ L_{\text{dis}} (f_{\psi_z}(x^{(i)}), z^{(i)}) +  L_{\text{dis}} (f_{\psi_s}(x^{(i)}), s^{(i)})
\end{align}

Since the learned shape code and deformation code in the first stage are disentangled, the inferred latent codes also remain disentangled. That means if two point clouds $x^{(i)}$ and $x^{(j)}$ are two different forms of the same shape, then they have the same shape code ($s^{(i)} = s^{(j)}$), but different deformation code ($z^{(i)} \neq z^{(j)}$). On the other hand, if two point clouds $x^{(i)}$ and $x^{(j)}$ are the same form of two different shapes, then they have the same deformation code ($z^{(i)} = z^{(j)}$), but different shape code ($s^{(i)} \neq s^{(j)}$). 

We used PointNet \cite{qi2017pointnet} to implement the encoder networks $f_{\psi_z}$ and $f_{\psi_s}$. The PointNet architecture deals with the permutation invariance of points in the input. We refer to the supplementary material for a detailed discussion of the PointNet architecture. 

\subsection{Implementation of Loss Functions}
To implement $L_\text{recon}(y^{(i)}, x^{(i)})$ in Eq. \ref{eq: latent} and Eq. \ref{eq: amortized}, we used pairwise Euclidean distances between all points in the 3D object.
\begin{align}
    L_\text{recon} (y^{(i)}, x^{(i)}) = \left\| \mathbf{D}_{\mathbb{R}^3}(y^{(i)}) - \mathbf{D}_{\mathbb{R}^3}(x^{(i)}) \right\|_F^2
\end{align}
Here, $\mathbf{D}_{\mathbb{R}^3}(x)$ is the matrix of pairwise Euclidean distances between all points
in $x$, and $\| \cdot \|_F$ denotes the Frobenius norm of the matrix. Such reconstruction loss has also been used in \cite{cosmo2020limp} and have been found to be highly effective for objects with same connectivity (as observed in SMPL, SMAL, COMA).

\section{Experiments \& Results}
\label{sec:result}

\begin{figure*}[t]
    \centering
    \includegraphics[width=1.0\linewidth]{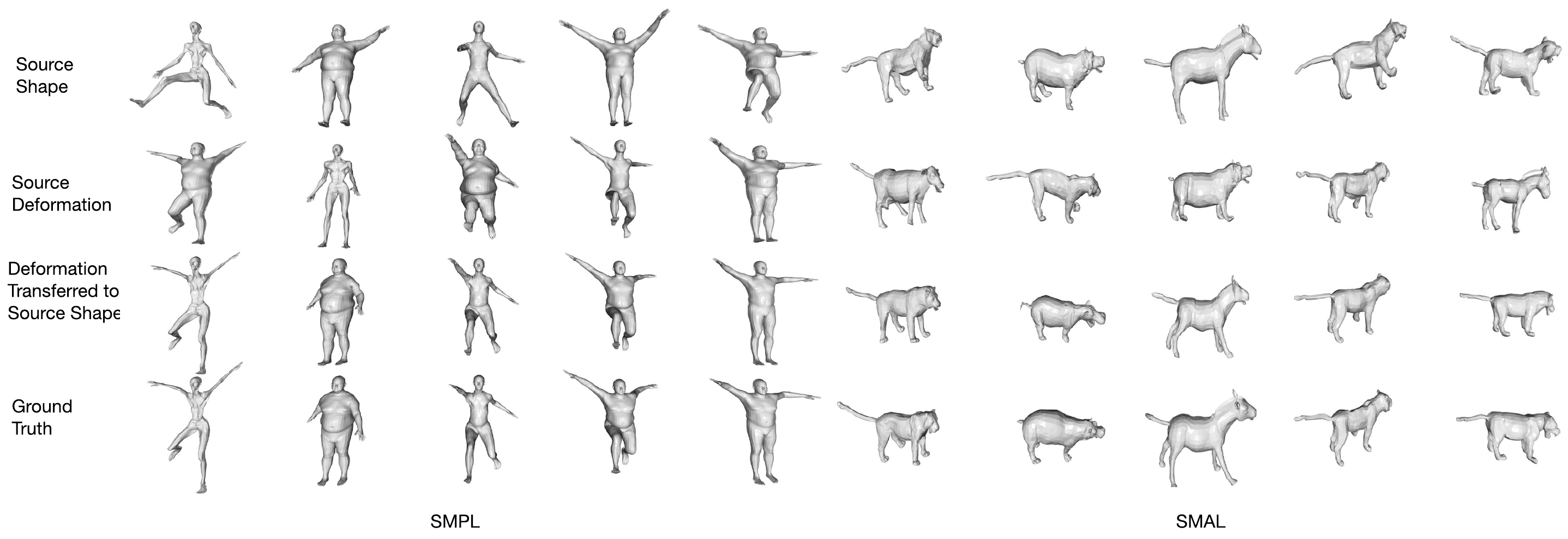}
    \caption{\textbf{Unsupervised 3D Deformation Transfer} in SMPL (left) and SMAL (right) datasets by our method. Additional visualizations can be found in the supplementary material.}
    \label{fig:pose-transfer}
\end{figure*}

\subsection{Implementation Details}
\label{sec: implementation}
We used PyTorch to implement our method and the baselines. We trained and tested them on NVIDIA RTX A5000 GPUs. We used the Adam optimizer with a cosine annealing learning rate scheduler to optimize the latent codes and networks. 
We used a batch size of 6 for SMPL and 16 for SMAL and COMA, which had fewer vertices. In all our experiments, the generator $ g_\theta$ used 5 AdaIn layers. For further implementation details and the code, we refer to the supplementary material.
\subsection{Datasets}
\textbf{SMPL-NPT: } This dataset \cite{wang2020neural} comprises 24,000 synthetic human meshes, each containing 6,890 vertices. It includes 30 shape identities represented in 800 unique deformations. For training, we draw 6,400 samples, covering 16 shapes and 400 deformations. For testing, following the literature \cite{song2023unsupervised}, we create a seen subset of 72 mesh pairs, sampled from the 14 shapes excluded from training and 400 deformations used in training. Similarly, we create an unseen subset of 72 mesh pairs, sampled from the remaining 14 shapes and 200 deformations which were excluded from the training data.\\
\textbf{SMAL: } This dataset \cite{zuffi20173d} contains 24,600 synthetic animal meshes, each containing 3,889 vertices. It represents 41 shape identities, each in 600 unique deformations. For training, we randomly sample 9,000 meshes from 29 shapes and 400 deformations. For evaluation, similar to the literature \cite{song2023unsupervised}, an independent set of 400 unseen mesh pairs is used, taken from the 12 shapes and 200 deformations that were excluded from training.\\
\textbf{COMA: } This dataset \cite{ranjan2018generating} contains meshes of 12 human faces, each performing 12 different facial expressions. By splitting some expressions into left and right variants, we obtained 17 distinct expressions in total. 10 subjects were used for training and 2 for testing, resulting in a test set with seen deformations but unseen identities. Since not all expressions are available for every subject, the training set contains 162 meshes and the test set 34 meshes. All meshes share the same connectivity, with 5023 vertices each.

\vspace{-0.5em}
\subsection{Evaluation Metrics}
We evaluated our method and the baseline methods in two different aspects:
\begin{enumerate}
    \item How effectively can the disentangled shape and deformation codes be applied for unsupervised deformation transfer on unseen (zero-shot) shapes or deformations?  
    \item To what extent can the disentangled shape and deformation codes predict actual deformations, and how independent are the factors in this prediction?
    
\end{enumerate}

\textbf{To assess criterion 1}, we utilize the evaluation metrics PMD and CD frequently used in the relevant literature. We define them as follows:

\noindent \textbf{PMD: }The average of the Euclidean distances between corresponding points in two point clouds.
\begin{align}
\label{eq: pmd_loss}
    L_\text{PMD} (y^{(i)}, x^{(i)}) = \frac{1}{N} \sum_{j=1}^{N} \| y_j^{(i)} - x_j^{(i)} \|^2
\end{align}
\noindent \textbf{CD: }Measures the similarity between two point clouds by calculating the average distance from each point in one set to the closest point in the other set. It is in the same form as Eq. \ref{eq: chamfer_loss}.
\begin{align}
\label{eq: chamfer_loss}
    L_\text{CD} (y^{(i)}, x^{(i)}) = M_{\alpha_c} \left( \frac{1}{|x^{(i)}|} \sum_{p \in x^{(i)}} \hat{d}(p), \frac{1}{|y^{(i)}|} \sum_{\hat{p} \in y^{(i)}} d(\hat{p}) \right)
\end{align}

\textbf{For criterion 2}, we define the predictivity and disentanglement score for deformation and use them as the evaluation metric. 

\begin{figure}[!ht]
    \centering
    \includegraphics[width=0.9\linewidth]{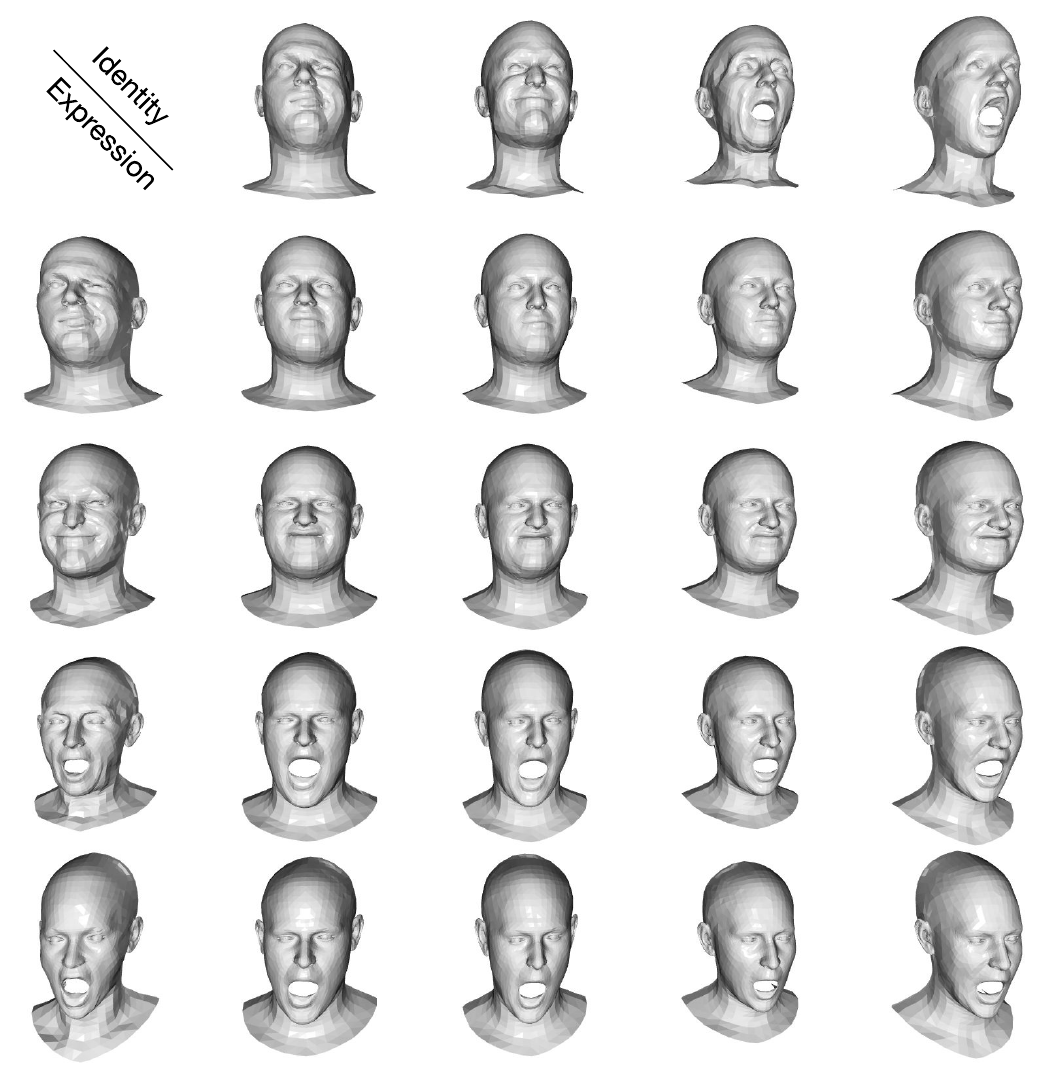}
    \caption{Qualitative results of DiLO on COMA. Top row: identity sources (shape codes). Left column: expression sources (content codes). Middle: generated faces combining identity and expression, accurately reflecting both source traits.}
    \label{fig:coma-transfer}
\end{figure}

\noindent \textbf{Latent Space Disentanglement Score: } Measuring disentanglement of shape and deformation in the latent space is inherently challenging. Nevertheless, a number of metrics, \textit{e.g.}, MIG score, SAP, $D_{\text{score}}$, etc., have been proposed to this end in the disentangled representation learning literature \cite{locatello2019challenging, detlefsen2019explicit}. Since \cite{locatello2019challenging} showed that these scores are highly correlated, using any one of them often suffices. Following the image style disentanglement methods \cite{detlefsen2019explicit, gabbaydemystifying}, we used disentanglement score $D_\text{score}$ \cite{detlefsen2019explicit} as the metric to measure disentanglement in latent space. If there are two latent codes $lc_1$ and $lc_2$, then the $D_\text{score}$ for any ground truth factor is defined as:
\begin{align}
    \label{eq: d_score}
    D_\text{score}(\text{factor}) = | E(\text{factor}|lc_1) - E(\text{factor}|lc_2)|
\end{align}
where $| |$ denotes absolute value and $E(\text{factor}|lc)$ means the predictivity of factor given only latent code $lc$. The predictivity is usually measured with a simple linear classifier. 
 
We further qualitatively assessed the efficacy of our method through explainability analysis using a surrogate model based explainability approach designed for 3D PointNet based neural networks. \cite{tan2022surrogate}.

\subsection{Unsupervised 3D Deformation Transfer} \label{sec: pose_transfer}
Disentangled shape and deformation codes in 3D generative models can be readily used for deformation transfer, making deformation transfer a natural way to evaluate disentanglement effectiveness. Given two 3D objects $x^{(s)}$ and $x^{(z)}$ where $x^{(s)}$ serves as the source shape and $x^{(z)}$ serves as the source deformation, the deformation transfer approach aims to predict the 3D object $x^{\text{new}}$ that has the shape of $x^{(s)}$ and deformation of $x^{(z)}$. Supervised 3D deformation transfer methods use the ground truth of deformation transferred object $x^{\text{new}}$ during training. On the other hand, unsupervised 3D deformation transfer methods do not use them for training. 

There exist two categories of unsupervised 3D deformation transfer methods- 1) unsupervised 3D methods directly predicting $x^{\text{new}}$ from the pair of objects $x^{(s)}$ and $x^{(z)}$ without disentanglement of latent codes and 2) unsupervised methods that disentangle shape and deformation code and can predict $x^{\text{new}}$ through latent manipulation and generation. In more detail, they infer the shape and deformation of latent codes with their encoder networks ($z=f_{\psi_z}(x^{(z)}), s=f_{\psi_s}(x^{(s)})$) and then predict $x^{\text{new}}$ using generator network by using codes $z$ and $s$ ($x^{\text{new}}=g_\theta(z,s)$).  

The most recent state-of-the-art methods in category 1 are X-DualNet \cite{song2023unsupervised} and MAPConNet \cite{sun2023mapconnet}. The existing unsupervised shape-deformation disentanglement methods leveraging the shape group information \cite{zhou2020unsupervised, cosmo2020limp, chen2021intrinsic} along with our method all belong to the category 2.

We have evaluated all these methods against the benchmark SMPL and SMAL datasets in unsupervised 3D deformation transfer experiments (Table \ref{tab:pose_transfer}). For the category 1 methods, we directly use their pretrained models to infer results on the benchmark test sets of SMPL and SMAL. For category 2 methods \cite{zhou2020unsupervised, cosmo2020limp, chen2021intrinsic}, we trained them ourselves on all the training datasets including COMA. However, we did not observe stable training with IEP-GAN \cite{chen2021intrinsic}, most likely due to its use of GANs. A similar phenomenon was also reported by \cite{song2023unsupervised}. Consequently, we exclude it from the performance comparison among the methods. We report the training time requirements of the category 2 methods in Table \ref{tab:time} which shows the sheer computational advantange of DiLO over other methods.

\begin{table}[h]
\centering
\resizebox{\linewidth}{!}{\begin{tabular}{l|c|c|c}
\hline
Method & Time/epoch  & Epochs trained & Total Training  \\
& (mins) ($\downarrow$) & trained & Time ($\downarrow$)\\
\hline
Zhou et al.   & 11.25                     & 100                     & 19 GPU hours                \\
LIMP           & 7.2                       & 200                     & 22 GPU hours                \\
DiLO           & \textbf{0.9}                       & 200                     & \textbf{3} GPU hours                 \\
\hline
\end{tabular}}
\caption{Computational cost comparison among DiLO and the related methods}
\label{tab:time}
\vspace{-1em}
\end{table}

\begin{table*}[!htb]
\centering
\resizebox{0.95\linewidth}{!}{\begin{tabular}{c|c|ccc}
\hline
\textbf{Dataset} & \textbf{Category} & \textbf{Method} & \textbf{PMD} ($10^{-3}$) ($\downarrow$) & \textbf{CD} ($10^{-3}$) ($\downarrow$) \\
\hline
SMPL & Deformation & X-DualNet \cite{song2023unsupervised} & 0.82 & 1.27\\
(unseen identities, & Transfer & MAPConNet \cite{sun2023mapconnet} & 0.52 & 1.02\\
\cline{2-5}
seen deforms) & \multirow{3}{*}{Disentanglement} & LIMP \cite{cosmo2020limp} & 20.21 & 32.24 \\
 & & Zhou et al. \cite{zhou2020unsupervised} &  \textbf{0.06} &  0.18\\
 & & Ours (w/o latent optimization) & 4.276 & 12.9 \\
 & & Ours (w/o AdaIn) & 5.86 & 21.40 \\
 & & Ours & \textbf{0.06} & \textbf{0.18} \\
\hline
SMPL & Deformation & X-DualNet \cite{song2023unsupervised} & 1.28 & 2.04\\
(unseen identities, & Transfer & MAPConNet \cite{sun2023mapconnet} & \textbf{0.74} & \textbf{1.45}\\
\cline{2-5}
unseen deforms)& \multirow{3}{*}{Disentanglement} & LIMP \cite{cosmo2020limp} & 26.38 & 43.64 \\
 & & Zhou et al. \cite{zhou2020unsupervised}  &  0.92 &  2.30\\
 & & Ours (w/o latent optimization) & 11.35 & 32.4 \\
 & & Ours (w/o AdaIn) & 13.93 & 37.44 \\
 & & Ours & 3.35 & 7.72 \\
\hline
\multirow{6}{*}{SMAL} & \multirow{2}{*}{Deformation Transfer} & X-DualNet \cite{song2023unsupervised} & 4.36 & 8.18\\
& & MAPConNet \cite{sun2023mapconnet}& 3.66 & \textbf{6.94}\\
\cline{2-5}
& \multirow{3}{*}{Disentanglement} &  LIMP \cite{cosmo2020limp} & 26.77 & 24.97 \\
& & Zhou et al. \cite{zhou2020unsupervised}  & 3.46 & 7.02\\
& & Ours (w/o latent optimization) & 9.66 & 23.6 \\
& & Ours (w/o AdaIn) & 6.61 & 13.14 \\
& & Ours & \textbf{3.45} &  7.40 \\
 \hline
\multirow{5}{*}{COMA} & \multirow{3}{*}{Disentanglement} & LIMP \cite{cosmo2020limp} & 7.39 & 21.227 \\
& & Zhou et al. \cite{zhou2020unsupervised}  & 8.82 & 18.76\\
& & Ours (w/o latent optimization) & 6.61 & 16.29 \\
& & Ours (w/o AdaIn) & 5.54 & 15.19 \\
& & Ours & \textbf{4.09} &  \textbf{13.29} \\
\hline
\end{tabular}}
\caption{\textbf{Comparison of unsupervised deformation transfer accuracy for different methods on SMPL and SMAL Datasets}. PMD and CD are in units of $10^{-3}$. ($\downarrow$) means lower values are better.}

\label{tab:pose_transfer}
\end{table*}

We conducted both qualitative and quantitative evaluations of unsupervised 3D deformation transfer on benchmark datasets. Quantitative results (Table \ref{tab:pose_transfer}) show that DiLO achieves performance comparable to or better than baseline methods, despite its simplicity and low computational cost. Although MAPConNet consistently yields the best Chamfer Distance (CD) on SMPL and SMAL, it requires deformation labels during training—unlike category 2 methods such as DiLO. Among category 2 methods, DiLO and the approach by \cite{zhou2020unsupervised} perform similarly on SMPL and SMAL, but DiLO significantly outperforms \cite{zhou2020unsupervised} on COMA. DiLO also consistently outperforms LIMP \cite{cosmo2020limp} across all datasets. Notably, both DiLO and LIMP use a simple encoder-decoder architecture with PointNet encoder, whereas \cite{zhou2020unsupervised} employ a complex multi-scale mesh-based encoder-decoder.

We provide a few representative qualitative results obtained with our method in Figure \ref{fig:pose-transfer} and Figure \ref{fig:coma-transfer}. More qualitative visualizations of our method and the baselines are available in the supplementary. Nevertheless, the figure demonstrates our method's successful unsupervised 3D deformation transfer without using any form of correspondence learning or ground truth target objects during training. 

\subsection{Disentanglement and Deformation Classification}
\label{sec: latent-space-disent}
To effectively measure disentanglement in the latent space, we assessed how predictive the latent codes are for the ground truth deformations not used in training. Thus we estimated the $D_\text{score} (\text{deformation})$ evaluation metric as described in \ref{eq: d_score}. 
\begin{table}[h!]
\centering
\renewcommand{\arraystretch}{1.5}
\resizebox{1.1\linewidth}{!}{\begin{tabular}{ccccc}
\hline
\textbf{Dataset} & \textbf{Method} & $\mathbf{E}(\text{Def.}|\mathbf{z})$ ($\uparrow$)& $\mathbf{E}(\text{Def.}|\mathbf{s})$ ($\downarrow$) & $\textbf{D}_\textbf{score}  (\text{Def.})$ ($\uparrow$)\\
\hline
\multirow{4}{*}{SMPL} & Zhou et al. & 0.918 & 0.085 & 0.833\\
 & LIMP &  0.960 &  0.940 & 0.020\\
& Ours (w/o LO) & 0.991 & 0.185 & 0.806\\
& Ours (w/o AdaIn) & 0.9175 & \textbf{0.000} & 0.9175\\
& Ours & \textbf{1.000} & 0.003 & \textbf{0.997}\\
\hline
 \multirow{4}{*}{SMAL} & Zhou et al.  & 0.718 & 0.010 & 0.708\\
 & LIMP & 0.390 & 0.323 & 0.067\\
 & Ours (w/o LO) & 0.725  &  0.018& 0.707\\
 & Ours (w/o AdaIn) & 0.933 & \textbf{0.003} & 0.930\\
 & Ours & \textbf{0.950} & 0.005 & \textbf{0.945}\\
 \hline
 \multirow{4}{*}{COMA} & Zhou et al. &  0.147&  0.118& 0.029\\
 & LIMP &  0.176&  0.088& 0.088\\
& Ours (w/o LO) & 0.235 & 0.118  & 0.117\\
 & Ours (w/o AdaIn) & 0.375 & 0.063 & 0.312\\
 & Ours & \textbf{0.656} & \textbf{0.063} & \textbf{0.593}\\
\hline
\end{tabular}}
\caption{Comparison of Deformation Prediction for Different Methods on SMPL, SMAL, and COMA Datasets. ($\uparrow$) indicates higher values are better. ($\downarrow$) means lower values are better. Def. and LO are used as an abbreviation for deformation factor and latent optimization, respectively.}
\label{tab:deformation_comparison}
\end{table}
We evaluated our method and other unsupervised shape deformation disentanglement methods on the SMPL and SMAL datasets. After estimating latent codes for both train and test datasets, we trained a linear support vector classifier (SVC) model with train latent codes and their labels. The model's prediction on the test dataset was used to report $\mathbf{E}(\text{Def.}|\mathbf{z})$ and $\mathbf{E}(\text{Def.}|\mathbf{s})$, with $\textbf{D}_\textbf{score} (\text{Def.})$ being the absolute difference between them. We did not report $D_\text{score} (\text{shape})$ since shape identities were already used during the generative model training in our method and the baseline methods. However, none used the deformation labels during training. It is important to note that the evaluation datasets for SMPL and SMAL used here differ from those in the deformation transfer experiments. Each evaluation dataset contains 400 random objects with unseen identities during training. For COMA, we use the same evaluation dataset used in deformation transfer experiment.

We report the estimated values in Table \ref{tab:deformation_comparison}. The table illustrates the superior latent space disentanglement achieved by our method compared to the baselines. It also demonstrates the efficacy of our method in accurately classifying deformation in 3D datasets using the inferred deformation latent code.

\begin{figure}[!ht]
\vspace{-1em}
    \centering
    \includegraphics[width=0.8\linewidth, height=0.9\linewidth]{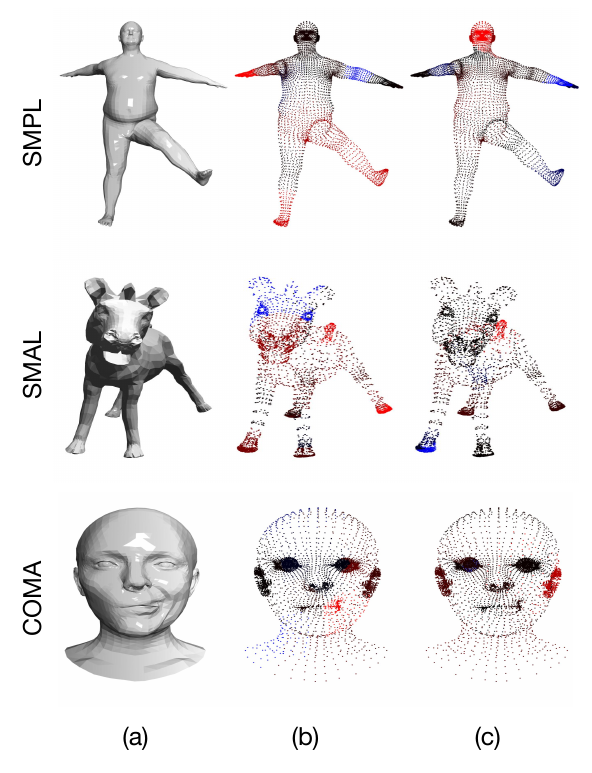}
    \caption{Results on explainability of DiLO. (a) A sample 3D mesh (b) Vertex importance learned by DiLO content encoder (c) Vertex importance learned by DiLO class encoder. Red represents high importance, blue represents low importance.}
    \label{fig:xai}
    \vspace{-1em}
\end{figure}

\subsection{Explainability Analyses}
\label{sec: xai}
We further performed explainability analysis of our model using LIME3D \cite{tan2022surrogate}. We assessed how the vertices at different location of the 3D objects in our datasets affects the shape encoder and the deformation encoder in DiLO. The detailed mechanism of how we leveraged LIME3D for this task is described in the supplementary document.  

We provide the explainability results on Figure \ref{fig:xai}. The figure shows that our deformation encoder reflects a high important to the vertices that gets affected by deformation. For instance, it focuses on the leg and hands in SMPL; leg, hand, and mouth in SMAL; and lips in COMA. On the other hand, our shape encoder assigns least importance to them. Instead, it prioritizes vertices associated with the identity of the 3D shapes- such as, face and body in SMPL, tail in SMAL, and ears in COMA. 

These results indicate that our method not only produces feasible outputs but also makes rational and interpretable choices throughout the learning process. This enables DiLO to serve as a valuable tool for investigating regions of interest in 3D object datasets. 

Furthermore, the explainability is a unique advantage provided by DiLO over the baseline methods. The baseline LIMP \cite{cosmo2020limp} has limited explainability due to its use of a single PointNet encoder for jointly predicting shape and deformation. On the other hand, \cite{zhou2020unsupervised} employs a multiscale mesh encoder-decoder architecture, where methods like LIME3D are not applicable. As a result, DiLO stands out from existing most relevant baselines by offering both strong performance, less computational cost, and practical explainability.

\noindent \textbf{Ablation Study: } We conducted ablation study to evaluate the contribution by individual components of DiLO, particularly 1) the latent optimization stage and 2) AdaIn layers in the generator. The ablative results provided in Table \ref{tab:pose_transfer} and \ref{tab:deformation_comparison} underscores the importance of two-stage training and using adaptive instance normalization in our final method.

\section{Discussions \& Conclusion}
In this work, we proposed a novel method termed DiLO for parameterizing deforming grouped 3D objects into disentangled shapes and deformation factors in an unsupervised manner. Similar to other existing effective methods for unsupervised 3D shape-deformation disentanglement, it uses the group information of the 3D objects and performs disentangled latent optimization. In most practical applications, such shape grouping information is readily available. We empirically demonstrate the importance of our two-stage framework and the overall design of our method through an extensive ablation study. Our successful demonstrations of unsupervised 3D deformation transfer, deformation classification, and explainability analysis position our method as a promising practical tool to advance unsupervised 3D vision.

\section{Acknowledgement}
This work was supported in part by the U.S. NIH grant R35GM158094.

\bibliography{main}
\newpage
\twocolumn[
\begin{center}
    {\LARGE \textbf{Supplementary material for DiLO:}\\
    \textbf{Disentangled Latent Optimization for\\
    Learning Shape and Deformation in Grouped\\
    Deforming 3D Objects}}
\end{center}
\vspace{1em}
]

\noindent \section{Related Works}
This supplementary section extends the Related Works section in the main manuscript. Here, we discuss the related works that we could not cover in our main manuscript due to page limit.

\textbf{Facial expression disentanglement: } There exists several works solely focusing on the facial expression disentanglement problem. For instance, \cite{zhang2020learning, olivier2023facetunegan} primarily targets solving the problem of facial expression disentanglement through adversarial training. \cite{zhang2020learning} uses a variational auto-encoder with Graph Convolutional Network (GCN), namely Mesh-Encoder, to model the distributions of identity and expression representations of 3D Faces via variational inference. To disentangle facial expression and identity, they eliminate correlation of the two distributions, and enforce them to be independent by adversarial training. \cite{olivier2023facetunegan} used an autoencoder with AdaIn layers for style transform. Additionally, they use a combination of reconstruction loss, adversarial loss, cycle-consistency loss, style reconstruction loss, feature matching loss, discriminator regularization loss, and laplacian smoothing loss to achieve disentanglement of facial identity and expression. Our work, on the other hand, does not depend on adversarial training and also does not specifically target only the facial expression disentanglement problem.

\textbf{Monocular depth sequence modeling: } There exists several methods \cite{palafox2021npms} that uses neural parametric models to simulate monocular depth sequences. Among these works, the closest to us is \cite{palafox2021npms} that uses latent optimization for fitting monocular depth sequences. Two different decoders, shape-decoder and pose-decoder, are used. A single shape code is used to learn the template shape in a monocular depth sequence. Pose codes are learned for each instance in the sequence. The pose decoder takes the shape code and pose code as inputs and outputs the deformation of a shape with respect to the template as a flow vector. Similar to this work, we use a single shape code across a group and a single pose code for an instance. However, we use a single decoder to generate the 3D instance from the shape and pose codes. We do not require any template to be given, nor do we require the deformation with respect to the template shape to be estimated for each instance. Moreover, unlike this work, our method concerns shape collections with multiple shape groups with intra-group variations and not single monocular depth sequences. 

\textbf{Disentangled representation learning for 2D images: } Learning disentangled representation for 2D images is a related and widely studied topic. While primitive methods like ICA and PCA also provided disentangled representation to some extent, the topic was popularized by the seminal works of $\beta$-VAE \cite{burgess2018understanding} and InfoGAN \cite{chen2016infogan}. Afterward, a series of works \cite{kim2018FactorVAE, chen2018betaTCVAE} were done to disentangle the generative factors of images using VAE-based methods. Works specifically designed to disentangle content and style of 2D images have also been conducted \cite{detlefsen2019explicit, gabbaydemystifying, gabbay2021scaling, richardson2021encoding}. These include methods based on VAEs, GANs, and latent optimization. Our work is highly motivated by the content-style disentanglement based on latent optimization. Nevertheless, we worked on the problem of shape deformation disentanglement in 3D objects represented using point clouds or meshes, which are very different from pixel-based 2D images.

\noindent \section{Implementation and Training Details}
In this supplementary section, we describe the implementation and training details, including the hyperparameter settings for our method and the baseline methods used in our experiments. This section elaborates Section \ref{sec: implementation} of the main manuscript with much more details.

We implemented our method and trained the baselines using PyTorch. We trained the models on NVIDIA RTX A5000 GPUs. 

\noindent \subsection{Implementing and Training DiLO}
While training our method-DiLO, in the first stage, only the generator and modulator network (that predicts the AdaIN parameters) were trained with the optimization of latent embeddings. The generator network consisted of $3$ fully connected linear layers followed by $5$ AdaIn layers, further followed by $2$ fully connected layers. The output feature dimensions of the first $3$ fully connected layers were $512$, $1024$, and $4096$. The output feature dimensions of the last $2$ fully connected layers were $64$ and $3$ (3 vertices of the point cloud). The $5$ AdaIn layers had feature dimensions of $16, 64, 256, 1024$, and $4096$, respectively. The initial learning rate for the training was fixed at $10^{-4}$ during this stage. The initial learning rate for optimizing the shape and latent codes was $3\times 10^{-3}$. A cosine annealing scheduler was used to update the learning rate until it reached a minimum value of $1\times 10^{-5}$.

\begin{table}[h!]
\centering
\renewcommand{\arraystretch}{1.5} 
\resizebox{\linewidth}{!}{\begin{tabular}{c|c|c|c}
\hline
\textbf{Hyperparamaters} & \textbf{SMPL-NPT} & \textbf{SMAL} & \textbf{COMA} \\ \hline
Shape dimension: & 256 & 256 & 256\\ 
Deformation dimension: & 128 & 128 & 128\\ 
No. epochs (first stage): & 300 & 200 & 200\\ 
Batch size (first stage): & 6 & 16 & 16\\ 
No. epochs (second stage): & 300 & 200 & 200\\ 
Batch size (second stage): & 6 & 16 & 16\\ \hline
\end{tabular}}
\caption{List of hyperparameter values for training DiLO against different datasets.}
\label{tab:hyperparam}
\end{table}

In the second stage of amortized inference, two distinct encoders were trained alongside the generator and modulator network. One encoder predicted the shape codes, and the other predicted the deformation codes. Both encoders were based on the PointNet architecture and were followed by fully connected layers to produce the final latent codes. The PointNet architecture consisted of $5$ layers of $1D$ convolutions, with each layer followed by Batch Normalization and ReLU non-linearity. Affine transformers were applied before the first and third convolutional layers. The feature dimensions for the convolution layers were fixed at $3, 50, 100, 200$, and $300$, respectively. $1D$ max pooling operation was used on the output of the last convolution layer. The fully connected layers mapped the pooled features to the embedding space. Batch Normalization and ReLU non-linearity were also applied here. The dimension of the fully connected layer feature was $500$.

In this stage, the initial learning rate for the generator and modulator was also fixed at $10^{-4}$. However, the initial learning rates for the encoders were fixed at $4\times 10^{-4}$. A cosine annealing scheduler was used to update the learning rate until it reached a minimum value of $1\times 10^{-5}$.

We put the remaining hyperparameter values used for training DiLO on different datasets in Table \ref{tab:hyperparam}.

\noindent \subsection{Training Zhou \emph{et al.} \cite{zhou2020unsupervised}}
This model used a Spiral Convolutional Auto-Encoder architecture to encode and decode shape and deformation information from mesh data. The encoder used spiral convolution layers for hierarchical feature extraction through downsampling, with four layers of increasing feature dimensions: [3, 4, 8, 16, 32] for shape and [3, 12, 24, 48, 96] for deformation. The flattened features were mapped to latent spaces of size 16 (shape) and 112 (deformation) via a fully connected layer. The decoder mirrored the encoder, using upsampling layers instead of downsampling, with feature dimensions [128, 64, 32, 16, 3]. Convolution parameters included hops [2, 2, 1, 1], dilation [1, 1, 1, 1], and scaling factors 4. Leaky ReLU with a negative slope of 0.02 was used as the activation function, and optimization was performed with the Adam optimizer and a cosine annealing learning rate scheduler starting at 1e-3. The training was conducted with batch sizes 6 for SMPL and 16 for SMAL and COMA datasets, using up/downsampling matrices derived from the reference meshes in the respective datasets.

\begin{figure*}[!htb]
    \centering
    {\includegraphics[width=0.9\linewidth]{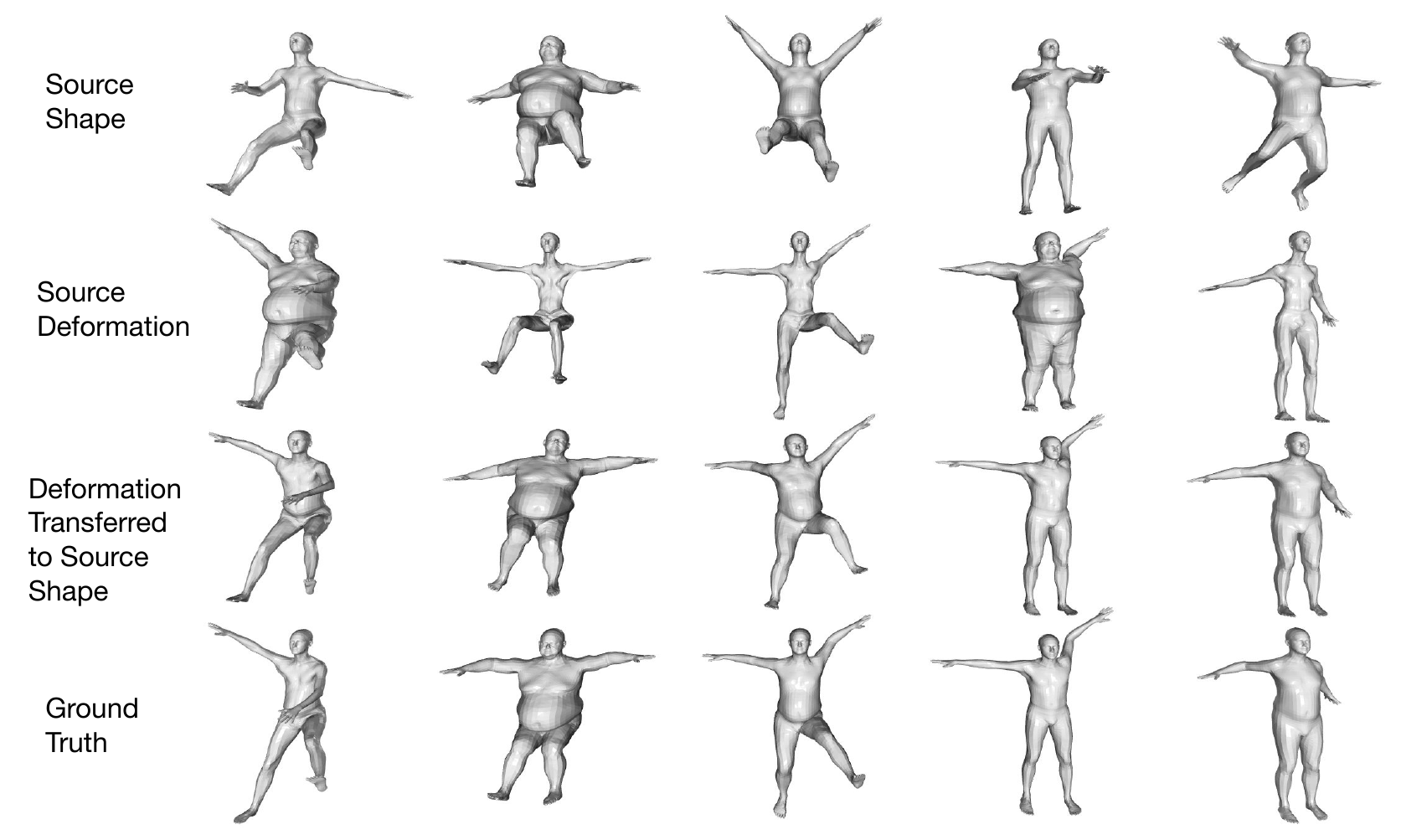}}
    \caption{Unsupervised 3D deformation transfer with DiLO on SMPL-NPT dataset}
    \label{fig:dilo-smpl}
\end{figure*}
\begin{figure*}[!htb]
    \centering
   {\includegraphics[width=0.9\linewidth]{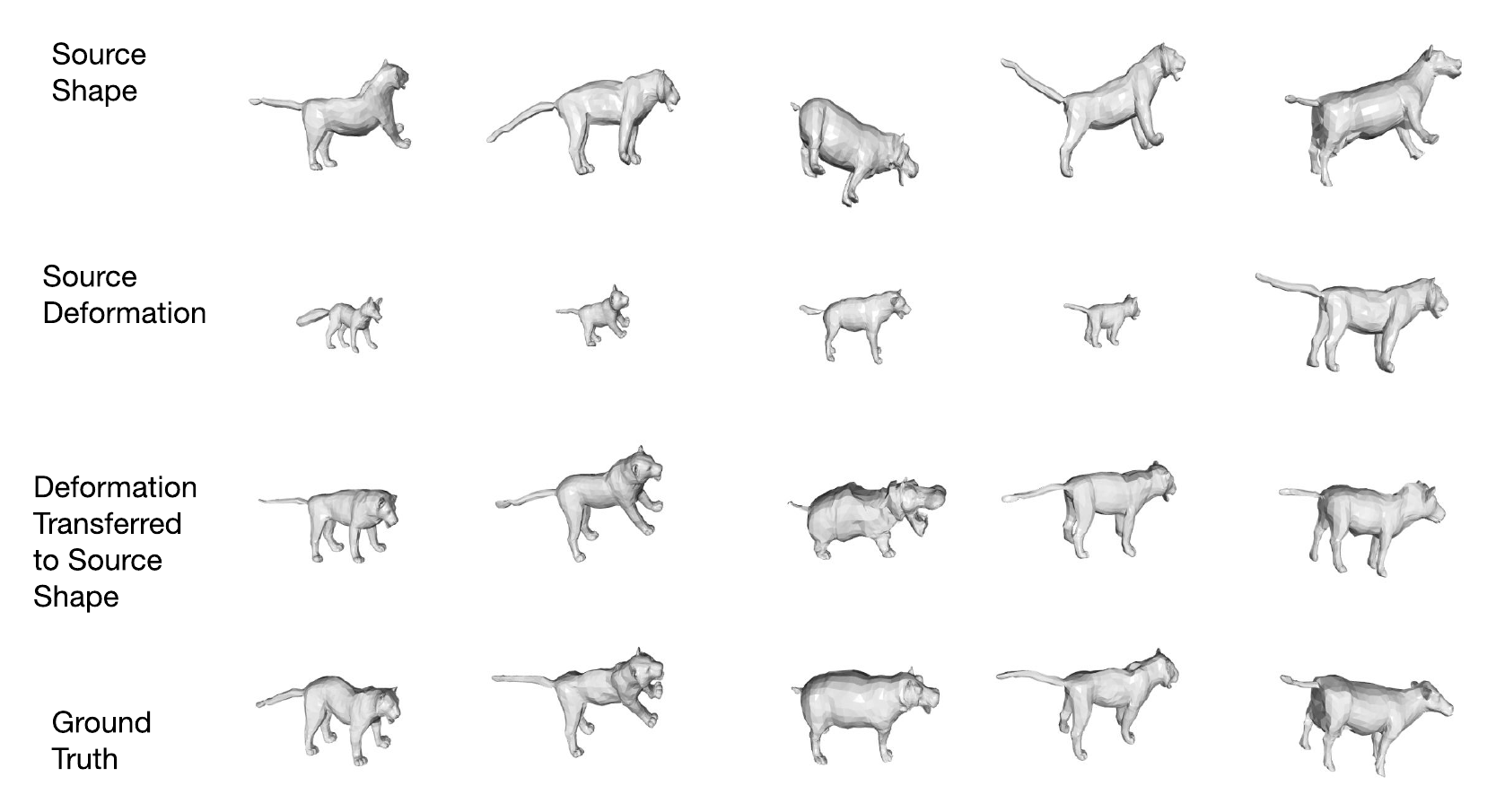}}
    \caption{Unsupervised 3D deformation transfer with DiLO on SMAL datasets.}
    \label{fig:dilo-smal}
\end{figure*}

\begin{figure*}[!htb]
    \centering
   {\includegraphics[width=0.8\linewidth]{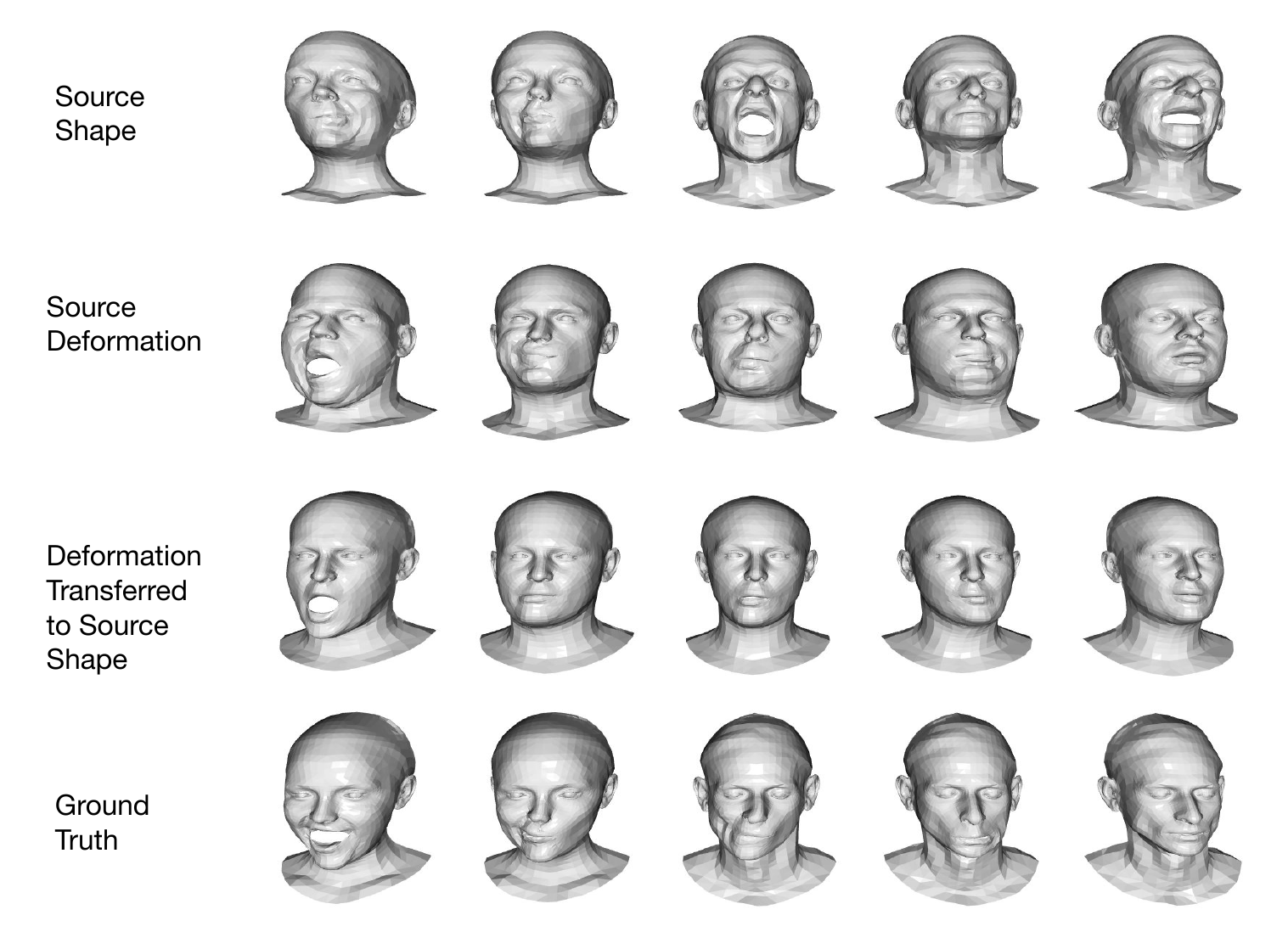}}
    \caption{Unsupervised 3D deformation transfer with DiLO on COMA datasets.}
    \label{fig:dilo-coma}
\end{figure*}

\begin{figure*}[!htb]
    \centering
   \includegraphics[width=0.8\linewidth]{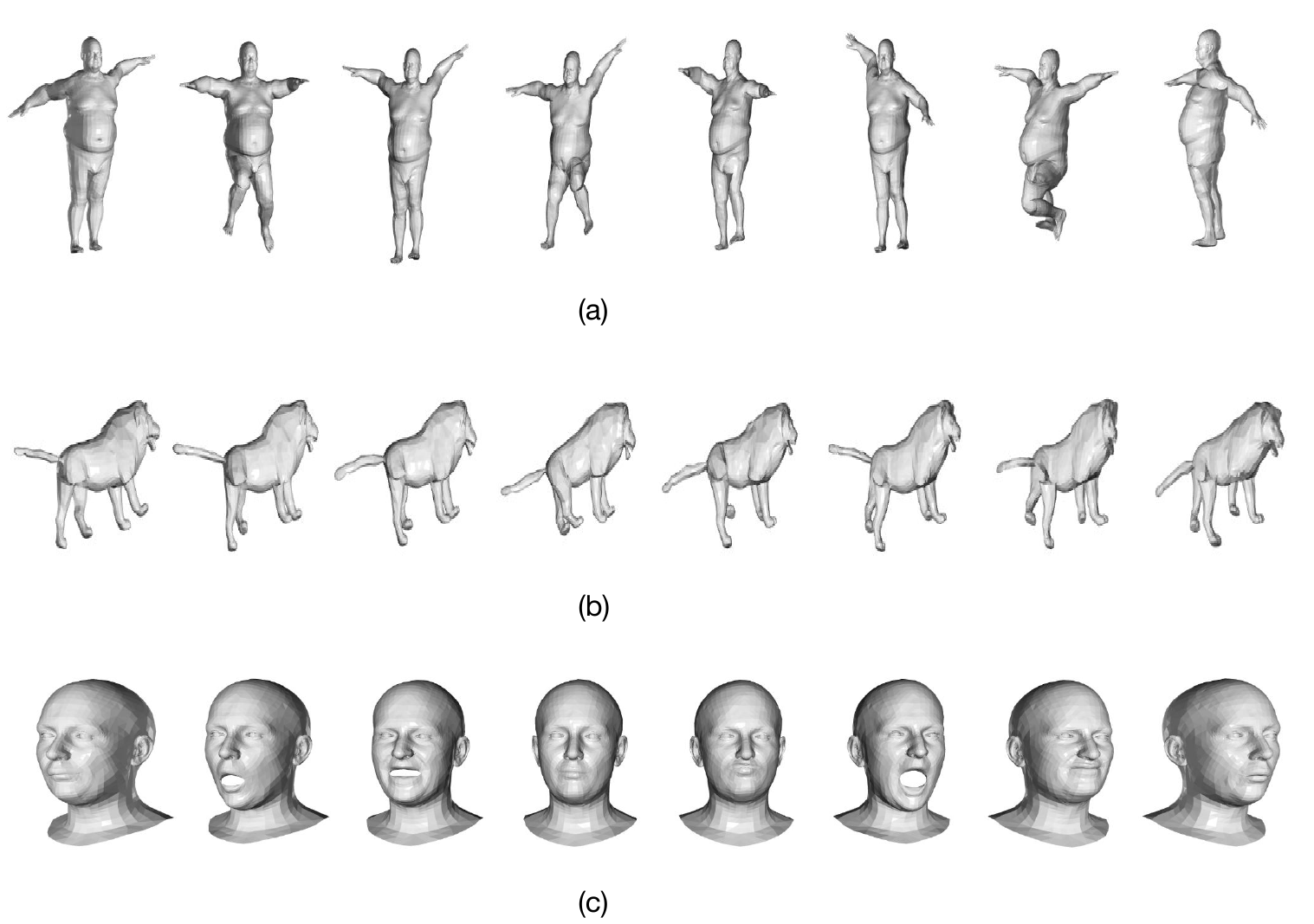}
    \caption{Generating 3D meshes with the same shape factor and different uniformly sampled deformation factors with DiLO on (a) SMPL-NPT, (b) SMAL, and (c) COMA datasets.}
    \label{fig:dilo-analysis}
\end{figure*}

\noindent \subsection{Training LIMP \cite{cosmo2020limp}}
LIMP uses a variational autoencoder (VAE) architecture to disentangle shape and deformation in the latent space. We trained LIMP on SMPL and SMAL datasets, mostly keeping the settings of their original code at \url{https://github.com/lcosmo/LIMP/tree/master}. The encoder was implemented using a PointNet encoder with an architecture similar to ours. The decoder was implemented using $3$ fully connected layers with output feature dimensions of $1024, 2048$, and $3$ (for 3 points in the point cloud), respectively. The first two fully connected layers were followed by leaky ReLU activation functions. The encoder outputs a single latent code of dimension $512$. The first $64$ dimensions were used as the deformation code, and the later $448$ dimensions were used as the shape code. The learning rate for training the model was set as $2\times 10^{-5}$. For shape-deformation disentanglement, LIMP requires the calculation of the geodesic distances of each object, which was performed using the Heat method of \cite{cosmo2016matching}. LIMP ensures that interpolating only the deformation latent code of different subjects preserves geodesic distances, and interpolating the shape latent code preserves the subject identity. For the preservation of geodesic distances, a threshold set is used as $0.1$. The model was trained for $20,000$ iterations, whereas for the first $14,000$ iterations, only the VAE reconstruction loss was used. From iterations $14,00$ to $16,000$, metric preserving interpolation and disentangling shape code loss were also used with the reconstruction loss. After $16,000$ iterations, disentangling deformation code loss with geodesic preservation was also employed in conjunction with the previous losses. This setting is similar to that used by \cite{cosmo2016matching}. We used a batch size of $4$. We used the same setting for SMPL-NPT, SMAL, and COMA datasets. 

\section{Additional Results}
This section extends the results section of the main manuscript by providing additional qualitative results on unsupervised 3D deformation transfer, latent space disentanglement, and explainability analyses with our method and other unsupervised shape-pose disentanglement methods. 

\subsection{Unsupervised 3D Deformation Transfer}
We provided a few sample qualitative 3D deformation transfer results for SMPL-NPT, SMAL, and COMA datasets with our method in the main manuscript. This section provides more 3D deformation transfer results with our method. 

\begin{figure*}[!htb]
    \centering
    \subfloat[\centering]{\includegraphics[width=0.8\linewidth]{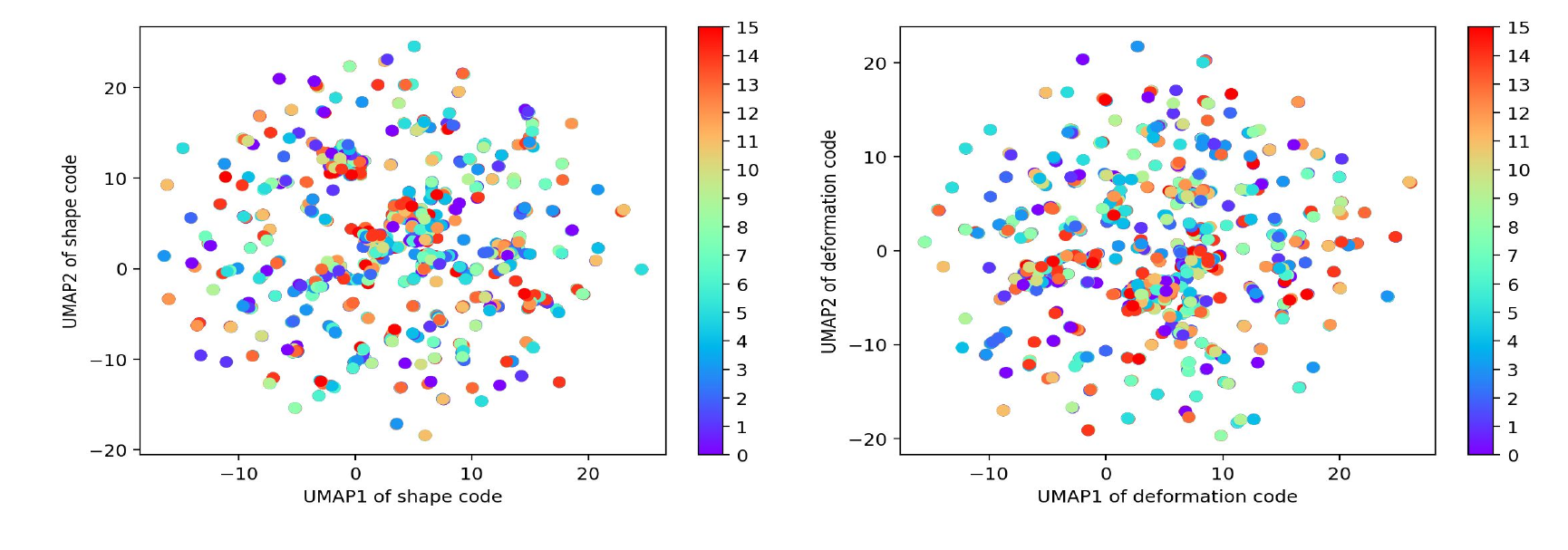}}
    \quad
    \subfloat[\centering]{\includegraphics[width=0.8\linewidth]{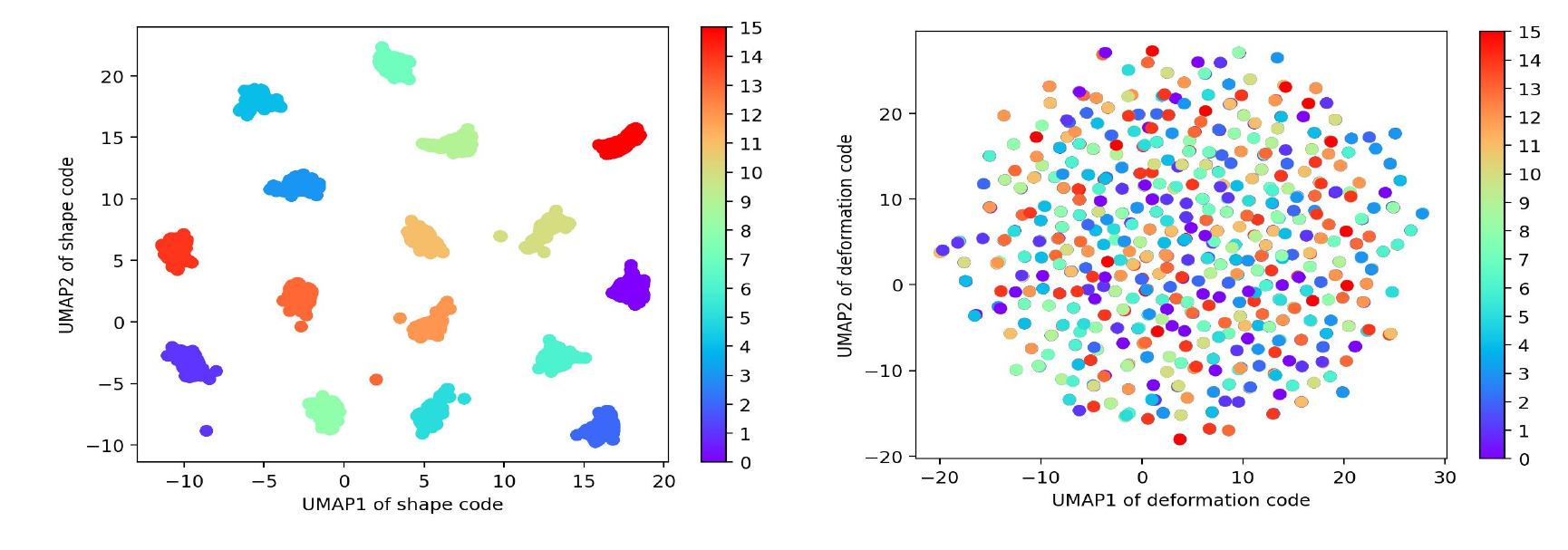}}
    \quad
    \subfloat[\centering]{\includegraphics[width=0.8\linewidth]{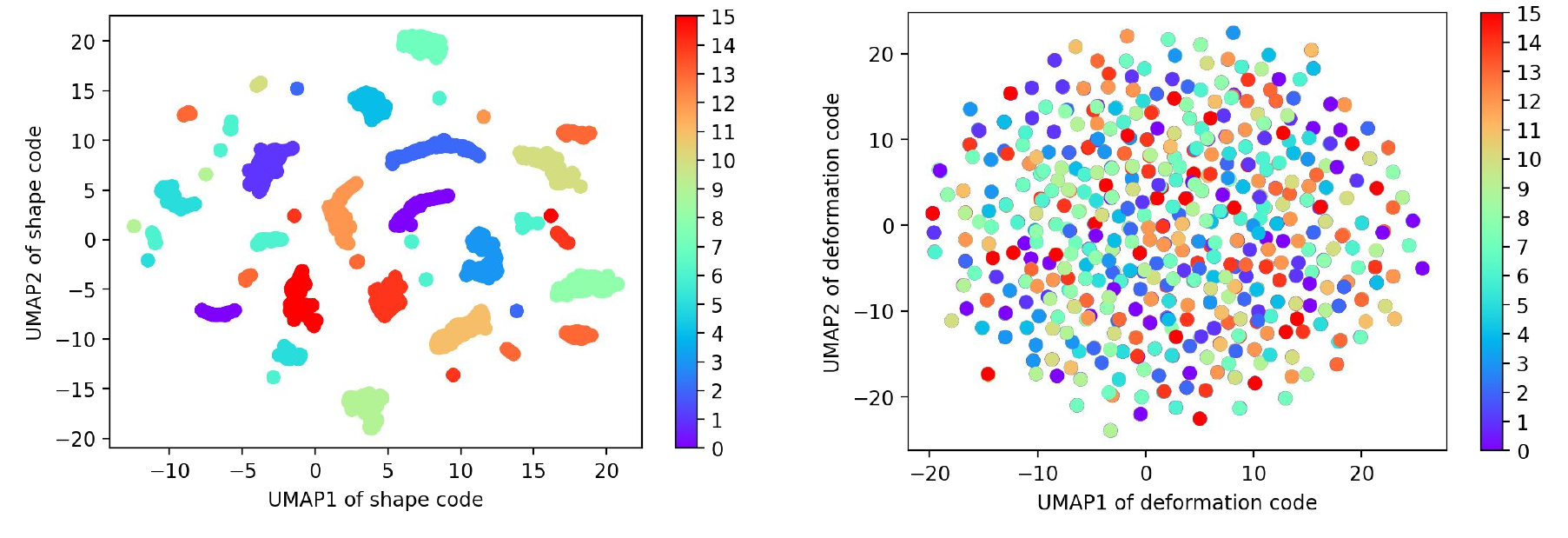}}
    \caption{Visualizing the UMAP of shape and deformation latent codes in (a) LIMP \cite{cosmo2020limp},  (b) Zhou et al. \cite{zhou2020unsupervised},  and (c) our method DiLO, for SMPL-NPT dataset. The samples are colored based on the subject identity (0-15).}
    \label{fig:smpl-clustering}
\end{figure*}

\begin{figure*}[!htb]
    \centering
    \subfloat[\centering]{\includegraphics[width=0.8\linewidth]{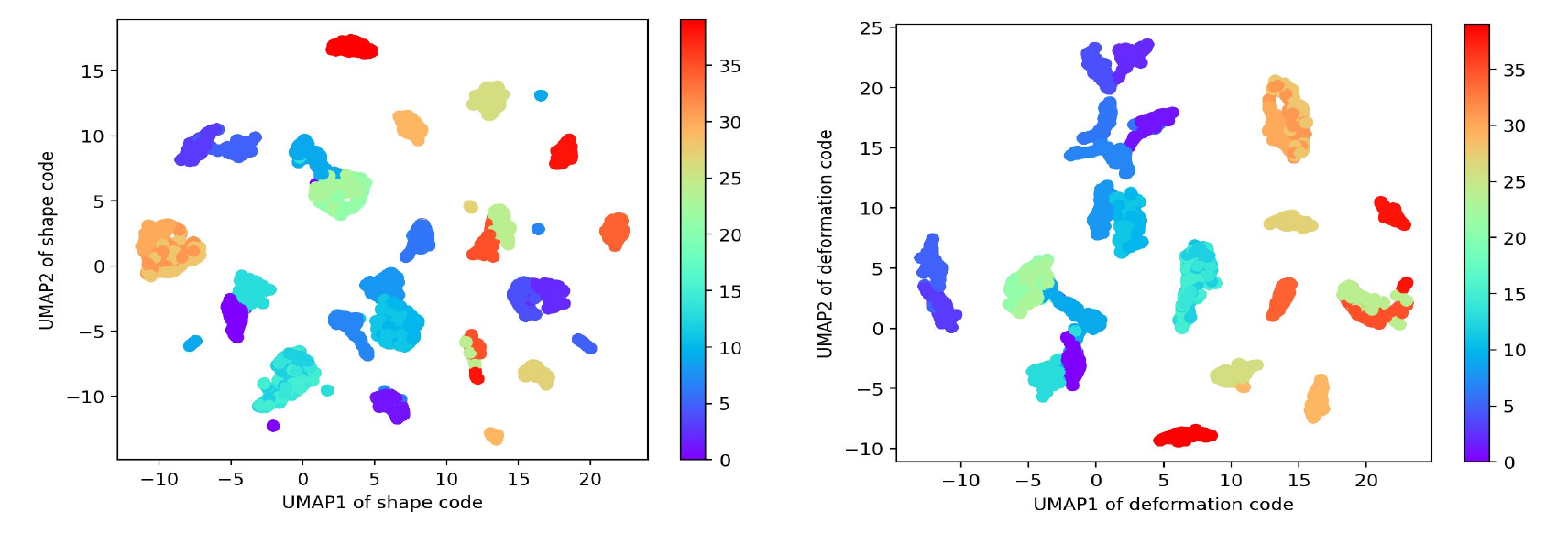}}
    \quad
    \subfloat[\centering]{\includegraphics[width=0.8\linewidth]{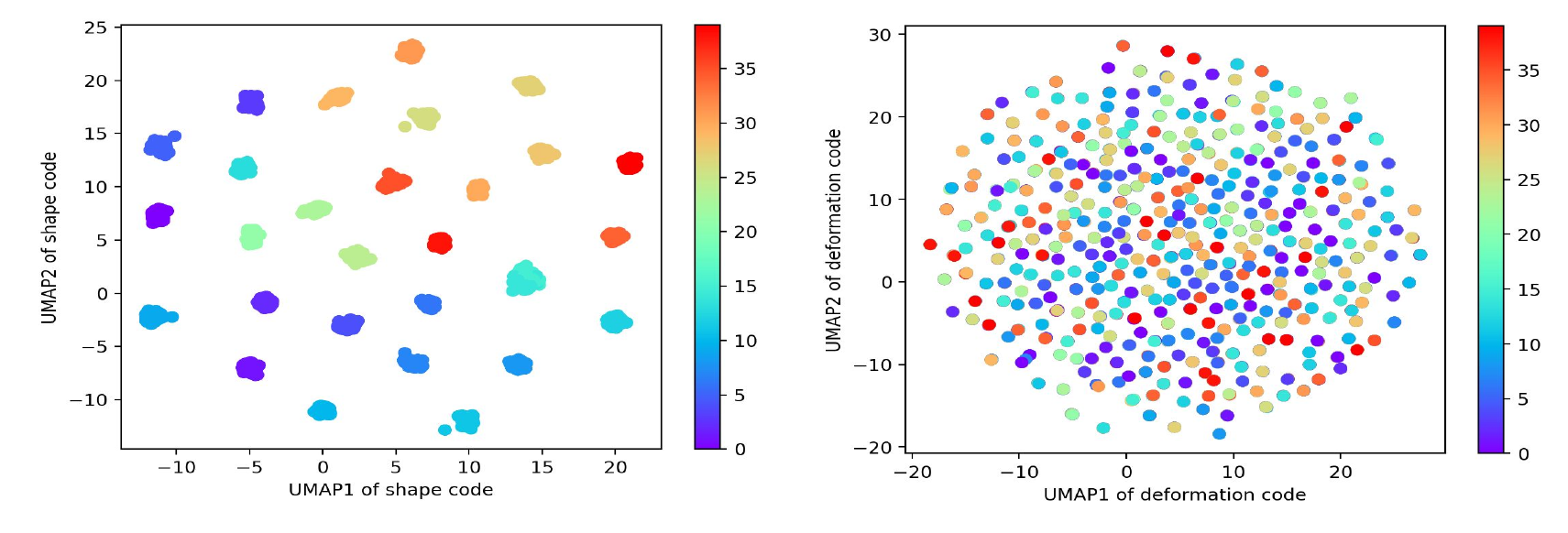}}
    \quad
    \subfloat[\centering]{\includegraphics[width=0.8\linewidth]{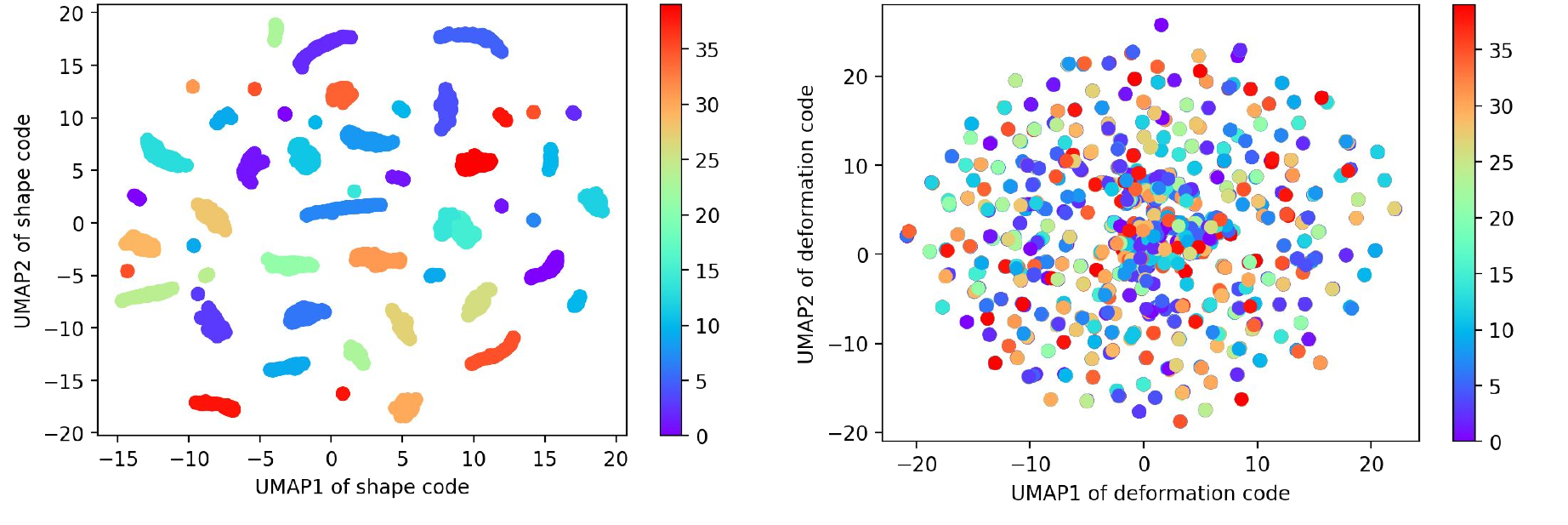}}
    \caption{Visualizing the UMAP of shape and deformation latent codes in (a) LIMP \cite{cosmo2020limp},  (b) Zhou et al. \cite{zhou2020unsupervised}, and (c) our method DiLO, for SMAL dataset. The samples are colored based on the subject identity (0-39).}
    \label{fig:smal-clustering}
\end{figure*}

\begin{figure*}[!htb]
    \centering
    \subfloat[\centering]{\includegraphics[width=0.8\linewidth]{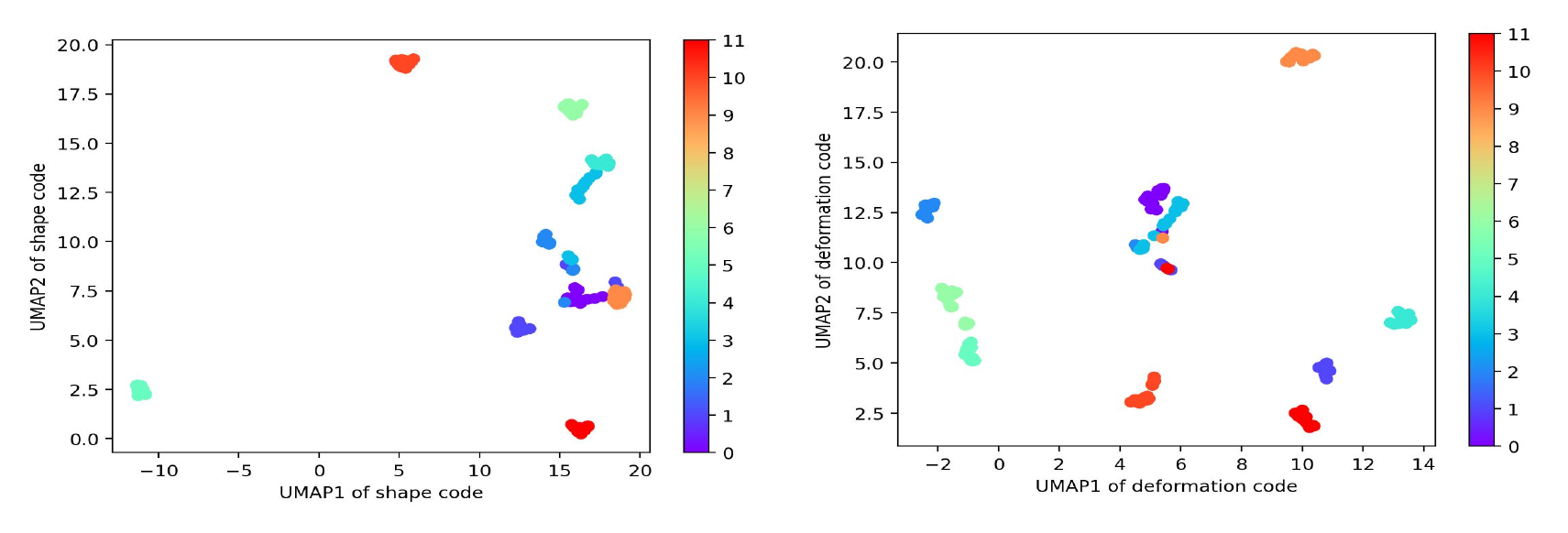}}
    \quad
    \subfloat[\centering]{\includegraphics[width=0.8\linewidth]{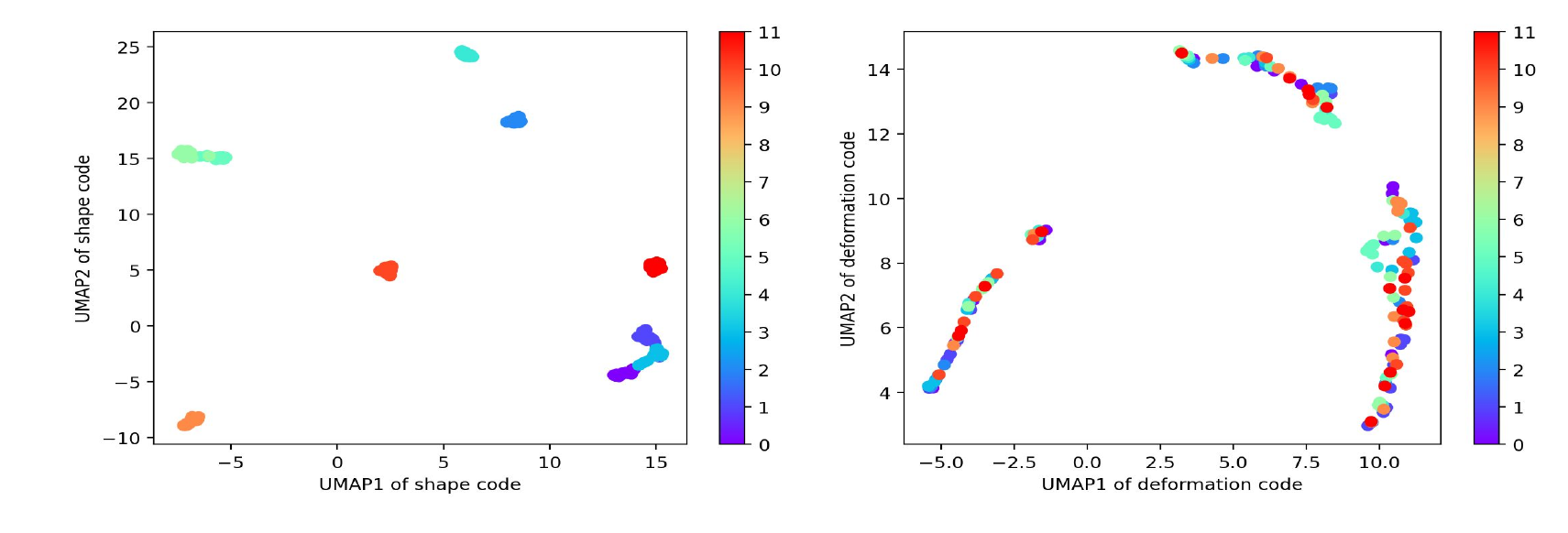}}
    \quad
    \subfloat[\centering]{\includegraphics[width=0.8\linewidth]{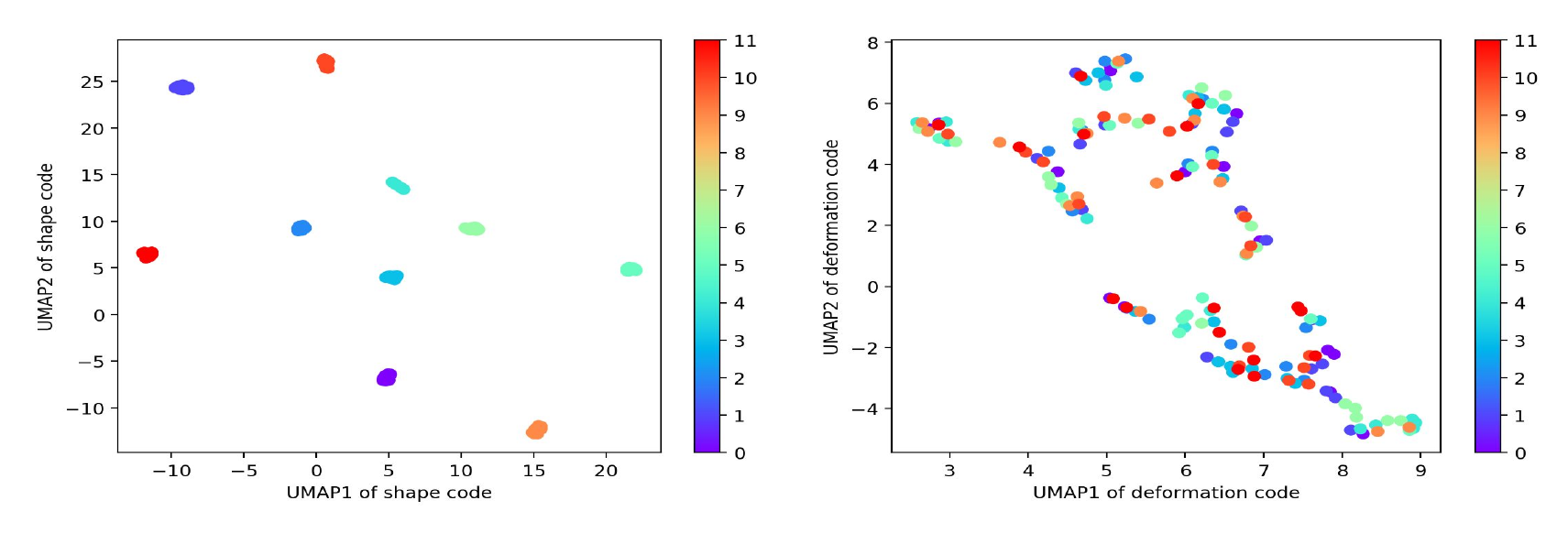}}
    \caption{Visualizing the UMAP of shape and deformation latent codes in (a) LIMP \cite{cosmo2020limp},  (b) Zhou et al. \cite{zhou2020unsupervised}, and (c) our method DiLO, for the COMA dataset. The samples are colored based on the subject identity (0-11).}
    \label{fig:coma-clustering}
\end{figure*}

We provide additional unsupervised deformation transfer results for SMPL, SMAL, and COMA datasets in Figure \ref{fig:dilo-smpl}, Figure \ref{fig:dilo-smal}, and Figure \ref{fig:dilo-coma} respectively. We further provide mesh generation results with DiLO with fixed shape code and varying deformation code across different datasets in Figure \ref{fig:dilo-analysis}.

\subsection{Latent Space Disentanglement}

We provided quantitative results on latent space disentanglement in Table 3 of the main manuscript. In this supplementary section, we provided qualitative results of latent space disentanglement through UMAP visualization. 

For our method DiLO and the baseline methods Zhou et al. and LIMP, we estimated 2 UMAP components for the shape latent codes and 2 UMAP components for the deformation latent codes. We did these for all of SMPL-NPT, SMAL, and COMA datasets. We plot the UMAP components for each object in the dataset, where we color the plot based on subject identity. We use subject identity (shape indicator) since the number of distinct subject identities is much smaller than the number of distinct deformation categories in SMPL-NPT, SMAL, and COMA datasets. In an ideal scenario, the UMAP of shape latent codes should clearly group the subject identities. In contrast, the UMAP of deformation latent codes should not exhibit any correlation with the subject identity grouping.   

We depict the UMAPs for our method DiLO, Zhou et al. and LIMP on the SMPL-NPT dataset in Figure \ref{fig:smpl-clustering}. Similarly, we provided the results for SMAL and COMA datasets in Figure \ref{fig:smal-clustering} and Figure \ref{fig:coma-clustering}.

Overall, for all the datasets, our method demonstrated excellent disentanglement of shape and deformation in its latent space. Zhou et al. also showed good disentanglement in the latent space. However, the disentanglement obtained by LIMP was largely limited in comparison.

\subsection{Explainability analysis}
For explainability analysis of our model, we used the surrogate model based LIME3D framework \cite{tan2022surrogate}. We assessed how the vertices at different location of the 3D objects in our datasets affects the shape encoder and the deformation encoder in DiLO.

\begin{figure*}[!htb]
    \centering
   \includegraphics[width=0.8\linewidth]{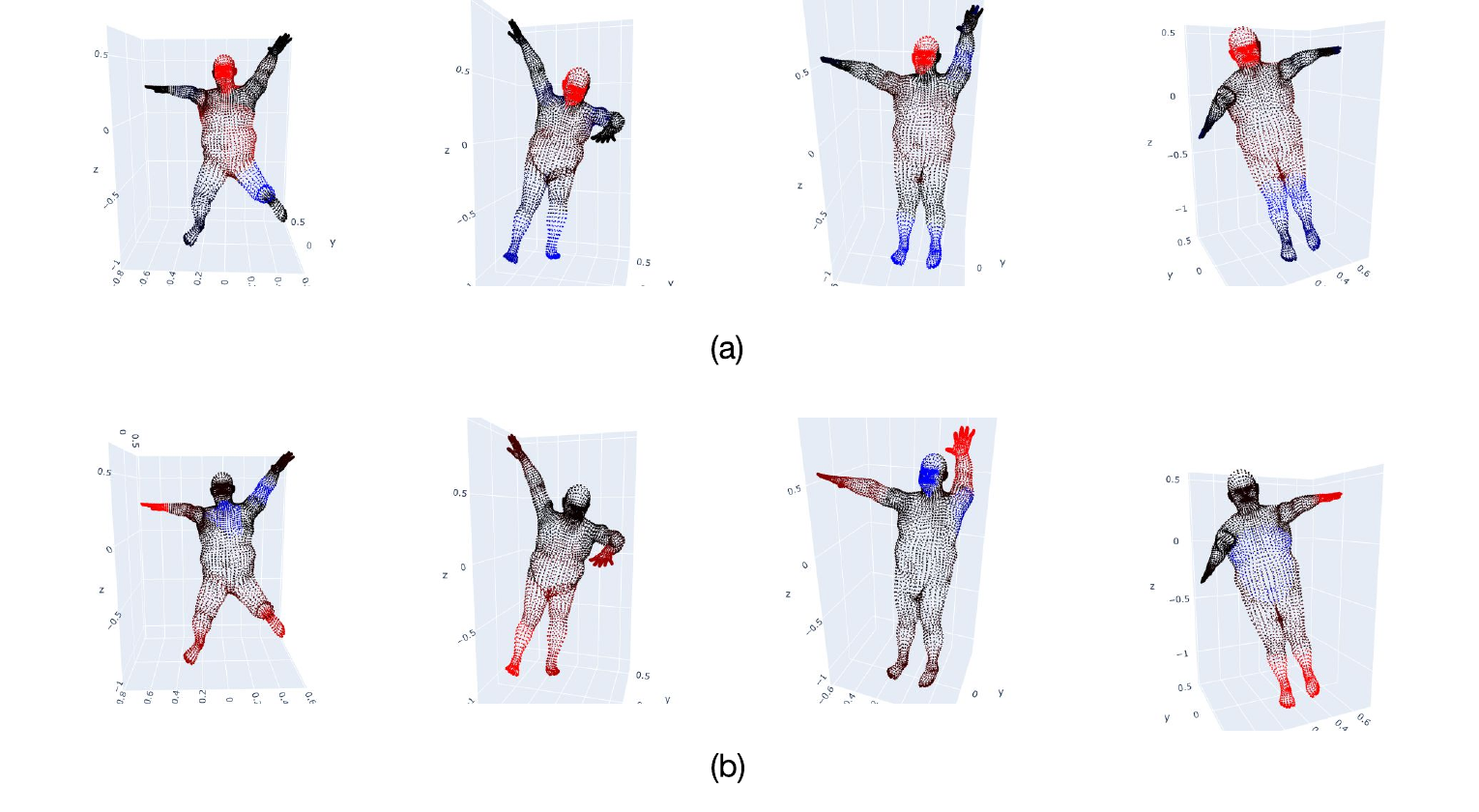}
    \caption{Importance of vertices of different 3D meshes to the (a) shape encoder and (b) deformation encoder of DiLO.}
    \label{fig:smpl-xai}
\end{figure*}

To this end, we first segmented the point cloud into spatial regions using KMeans clustering algorithm, treating each cluster as a discrete interpretable unit. We then generated binary masks by randomly selecting subsets of clusters to retain or mask out, effectively creating perturbed versions of the original input.

Each perturbed point cloud was passed to either the shape encoder or the deformation encoder to compute a model output. These outputs, along with their corresponding binary masks, were used to train a locally weighted linear regression model, where the weight for each sample was based on its cosine similarity to the original (unmasked) input.

The regression coefficients from this surrogate model quantified the contribution of each cluster to the prediction. Finally, we visualized the point cloud by coloring each point according to the contribution of its corresponding cluster—positive contributions were shown in blue, negative in red—thereby revealing which regions most influenced the model’s behavior.

In Figure 4 of the main manuscript, we showcase explainability results across three representative objects—one each from the SMPL, SMAL, and COMA datasets. We demonstrated how the shape encoder and the deformation encoder trained in DiLO is actually making rational choices while assigning importance to the vertices of the 3D meshes. In Figure \ref{fig:smpl-xai}, we further showed several examples of the importance assignment by the DiLO shape encoder and deformation encoder to various 3D shapes from SMPL dataset. We can observe that the shape encoder consistently focuses on the face and body shape, whereas the deformation encoder focuses on the hands and legs accountable for the deformation of the human objects. These results further strengthen the explainability aspect of our method.

\end{document}